\documentclass[10pt,journal,compsoc]{IEEEtran}
\ifCLASSOPTIONcompsoc
  \usepackage[nocompress]{cite}
\else
  \usepackage{cite}
\fi
\ifCLASSINFOpdf

\usepackage[ruled]{algorithm2e}
\usepackage{graphicx}
\usepackage{subfigure}
\usepackage{booktabs}
\usepackage{amsmath}
\usepackage[cmintegrals]{newtxmath}
\usepackage{multirow}
\usepackage{color}
\else
\fi
\usepackage{url}

\begin{document}

\title{OPOM: Customized Invisible Cloak towards Face Privacy Protection}

\author{Yaoyao Zhong,
	Weihong Deng
	\thanks{The authors are with Beijing University of Posts and Telecommunications, Beijing, China (e-mail: zhongyaoyao@bupt.edu.cn; whdeng@bupt.edu.cn).}
	\thanks{Datasets and code are available at https://github.com/zhongyy/OPOM.}
}

\IEEEtitleabstractindextext{%
\begin{abstract}
	While convenient in daily life, face recognition technologies also raise privacy concerns for regular users on the social media since they could be used to analyze face images and videos, efficiently and surreptitiously without any security restrictions. In this paper, we investigate the face privacy protection from a technology standpoint based on a new type of customized cloak, which can be applied to all the images of a regular user, to prevent malicious face recognition systems from uncovering their identity. Specifically, we propose a new method, named one person one mask (OPOM), to generate person-specific (class-wise) universal masks by optimizing each training sample in the direction away from the feature subspace of the source identity. To make full use of the limited training images, we investigate several modeling methods, including affine hulls, class centers and convex hulls, to obtain a better description of the feature subspace of source identities. The effectiveness of the proposed method is evaluated on both common and celebrity datasets against black-box face recognition models with different loss functions and network architectures. In addition, we discuss the advantages and potential problems of the proposed method. In particular, we conduct an application study on the privacy protection of a video dataset, Sherlock, to demonstrate the potential practical usage of the proposed method.        
\end{abstract}

}

\maketitle

\IEEEdisplaynontitleabstractindextext

\IEEEpeerreviewmaketitle

\IEEEraisesectionheading{\section{Introduction}\label{sec:introduction}}

\IEEEPARstart{D}{eep} learning has achieved considerable success in computer vision~\cite{k2012imagenet,simonyan2014very,he2016deep,hu2017squeeze}, significantly improving the state-of-art of face recognition~\cite{sun2014deep,Schroff2015FaceNet,wen2016discriminative,liu2017sphereface,chen2017noisy,Wang2018CosFace,deng2019arcface,huang2020curricularface,zhong2021sface}. This ubiquitous technology is now used to create innovative applications for entertainment and commercial services. 

However, face recognition can be used for good as well as ill. Due to privacy concerns and fears of an Orwellian surveillance society~\cite{bowyer2004face}, exploiting and analyzing face images without restriction has promoted considerable controversy. Unfortunately, such concerns and fears are not without merit. Human behavioral and psychological traits on social media and video surveillance, including photographs and videos, are at risk of unauthorized access or can be used inappropriately or with unintended disclosure. Privacy laws can be used to regulate the misuse of face recognition systems~\cite{IBMquit,ConvenienceAndPrivacyCollide} to avoid the potential risk of inappropriate usage of private information. However, simply prohibiting face recognition technique due to privacy concerns is not the best way. It may be better to develop deep learning technologies to solve the problems that the technology itself brings~\cite{newton2005preserving}. 

To protect privacy while maintaining practically usage, some studies~\cite{newton2005preserving,gross2009face,meden2017face,gafni2019live,li2019anonymousnet} aim to de-identify face images such that many facial characteristics remain, but the identities of people in the images cannot be reliably recognized. Considering that the generated images may have different visual appearances compared with the original images~\cite{newton2005preserving,gross2009face}, or behave unnaturally and exhibit undesirable artifacts~\cite{meden2017face,gafni2019live,li2019anonymousnet}, some recent methods~\cite{joon2017adversarial,chatzikyriakidis2019adversarial,shan2020fawkes,yang2020towards,zhang2020adversarial,cherepanova2020lowkey} can both hide the identification information and maintain the visual quality of face images, by generating imperceptible adversarial examples~\cite{Szegedy14} as privacy protection masks. Despite the naturalness and effectiveness, it is incredibly unfriendly for regular users to generate different privacy masks for each photograph or each frame of videos. 

Further study is necessary to provide more effective and simple face privacy protections for the general public. In contrast to existing methods, which generate different adversarial masks for the different face images of a person, we aim to generate a type of person-specific (class-wise) universal mask. In this way, a regular user can generate one privacy mask only once, then apply it to all his or her photographs and videos. 

Compared with image-specific privacy masks, the benefits of person-specific universal masks are twofold. First, the person-specific mask is generated once, therefore it can dispense with the mask generation time for new images, which may benefit average users and some real-time privacy protection applications in terms of efficiency. Second, compared with image-specific privacy masks which require multiple transmissions of new images between the user and the server, person-specific masks need only one transmission, which can reduce the risk of privacy leakage.  

There are two challenges in generating person-specific privacy masks.
(1) Individual Universality. Compared with image-specific privacy masks~\cite{joon2017adversarial,chatzikyriakidis2019adversarial,shan2020fawkes,yang2020towards,zhang2020adversarial,cherepanova2020lowkey}, person-specific privacy masks are generated with only a small number of images of an identity, and they can be applied to different unknown images. While face images of the same identity can vary from poses, illuminations, expressions and occlusions~\cite{cao2018vggface2}. This diversity will undoubtedly increase the difficulty of generating privacy masks for different face images. Therefore, it is crucial to increase the individual universality of adversarial masks for person-specific privacy protection.
(2) Model Transferability. The privacy masks should be transferable towards different models~\cite{liu2016delving,papernot2017practical,dong2018boosting,xie2019improving}, which means that they are generated from the local surrogate models and applied to different unknown recognition systems. For unknown face recognition models, there are a wide range of options for training databases, training loss functions and network architectures~\cite{zhong2020towards}. This variety will undoubtedly increase the difficulty of generating transferable adversarial masks.

\begin{figure*}[htbp]
	\center
	\includegraphics[width=1\linewidth]{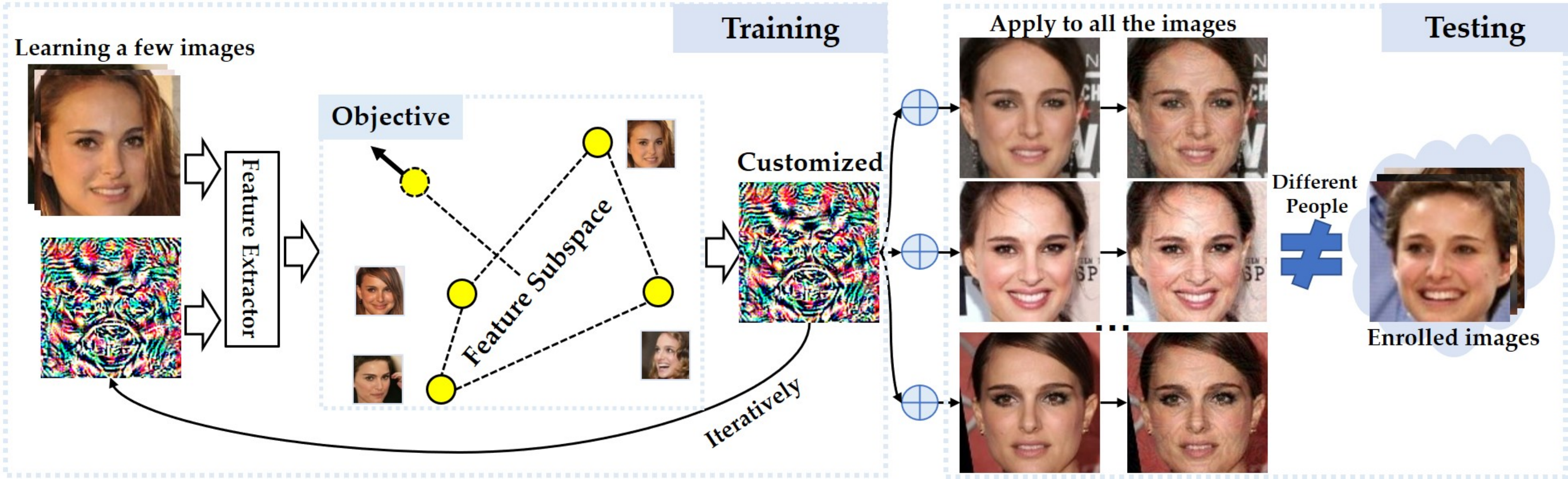}
	\caption{Illustration of person-specific (class-wise) universal privacy masks. A regular user can generate one privacy mask only once, and then apply it to all his or her photographs and videos. In the training process for mask generation, the proposed OPOM method optimizes each training sample in the direction away from the feature subspace of the source identity. To make full use of the limited training images, several modeling methods, including affine hulls, class centers and convex hulls, have been investigated to obtain a better description of the feature subspace of the source identities. In the testing process, with the customized mask, the protected images will not be recognized as the original identity. Compared with image-specific protection, person-specific masks protect the privacy of regular users in a friendlier way.}
	\label{fig:intro}
\end{figure*}

\begin{figure}[h]
	\center
	\includegraphics[width=1\linewidth]{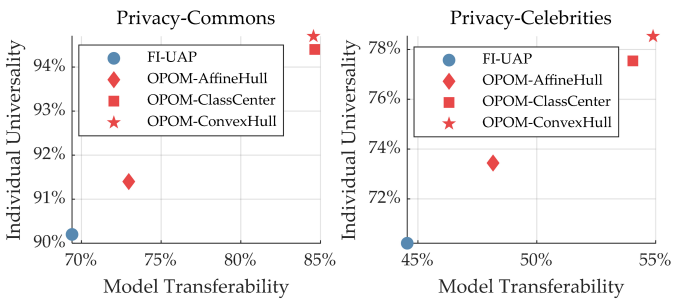}
	\caption{With better approximation methods (red) for the feature subspace of each identity in the mask generation process, both individual universality and model transferability can be improved, which will lead to more effective privacy protection. Experimental results on two datasets (Privacy-Commons and Privacy-Celebrities) are shown.}
	\label{fig:intro_toy}
\end{figure}

In this paper, we propose a method, named one person one mask (OPOM), to provide one privacy mask, for all the face images of one person, similar to a customized invisible cloak, as shown in Figure~\ref{fig:intro}. Specifically, to increase the individual universality, OPOM generates a privacy mask by solving an optimization problem, which maximizes the distance between diverse deep features of the training images and the feature subspace of the identity. We investigate different modeling methods, including affine hulls, class centers and convex hulls, to model the feature subspace of each identity for better description with limited images. As shown in Figure~\ref{fig:intro_toy}, we empirically find that, with better description of the feature subspace in the mask generation process, both individual universality and model transferability can be improved. Furthermore, to increase model transferability, OPOM is also designed to be combined with a variety of model transferability methods, such as the momentum boosting method~\cite{dong2018boosting} and DFANet~\cite{zhong2020towards}. The main contributions of our paper are as follows: 

\begin{itemize}
	\item{We reveal the existence of a new type of person-specific (class-wise) universal adversarial privacy mask, which is generated to protect different face images of the same identity, and therefore can be easier-to-use for regular users.}
	
	\item{We investigate this new type of person-specific privacy mask, and propose an efficient method, OPOM, to generate this type of adversarial mask, which can jointly improve the \emph{Image Universality} and \emph{Model Transferability}, therefore attaining more effective privacy protection.}
	
	\item{The effectiveness of the proposed OPOM method is empirically demonstrated in protecting unconstrained face images towards different black-box deep face recognition models compared with previous universal adversarial perturbations.}
\end{itemize}      

The remainder of the paper is organized as follows. Section~\ref{sec:related} briefly reviews the literature related to privacy protection and adversarial attacks. In Section~\ref{sec:method}, we first introduce the proposed OPOM for generating privacy masks; and then describe the potential comparable methods. In Section~\ref{sec:experiment}, we first evaluate the effectiveness of the proposed method; and then discuss the proposed method for a deeper understanding. Finally, Section~\ref{sec:conclusion} summarizes this paper and provides suggestions for future works. 


\section{Related Works}
\label{sec:related}
In this section, we briefly review the literature related to face privacy protection and adversarial attacks.

\subsection{Face privacy protection}
Privacy protection has been studied for a long time in the literature~\cite{cho1999privacy}, and has attracted more attention in the current deep learning era, since multimedia data can be analyzed accurately and efficiently with high-performance deep models~\cite{liu2021machine}. The face image, as an important type of biometric data, can reveal a large amount of identity information. 

With the help of deep learning technologies, face recognition has developed with unprecedented success~\cite{sun2014deep,Schroff2015FaceNet,wen2016discriminative,liu2017sphereface,chen2017noisy,Wang2018CosFace,deng2019arcface,huang2020curricularface,zhong2021sface}. Face recognition models are trained on large-scale training databases~\cite{Yi2014CASIA,cao2018vggface2,Guo16MS}, and used as feature extractors to test identities that are usually disjoint from the training set~\cite{liu2017sphereface}. With these open-set face recognition models~\cite{WANG2021215}, images and short videos posted casually on the social media can be collected and analyzed by unauthorized commercial services. 

To protect privacy, studies~\cite{newton2005preserving,gross2009face,meden2017face,gafni2019live,li2019anonymousnet} on face de-identification aim to remove identity information from the images. Newton \emph{et al}.\ \cite{newton2005preserving} replaced the original images with the average of $k$ face images, and Gross \emph{et al}.\ \cite{gross2009face} reconstructed a de-identified face using a generative multi-factor model, which were early attempts to face de-identification to replace masking, blurring, or pixelation techniques. In recent years, motivated by generative neural networks, face de-identification works~\cite{meden2017face,gafni2019live,li2019anonymousnet} have developed the synthesis images to replace the original images for privacy protection. Despite their effectiveness, considering the lack of naturalness of replaced faces in previous methods~\cite{newton2005preserving,gross2009face,meden2017face,gafni2019live,li2019anonymousnet}, some researchers~\cite{joon2017adversarial,chatzikyriakidis2019adversarial,shan2020fawkes,yang2020towards,zhang2020adversarial,cherepanova2020lowkey} are inspired by adversarial examples~\cite{Szegedy14} and made some successful attempts at face adversarial privacy protection.

\subsection{Adversarial Attacks}
Szegedy \emph{et al}.\ \cite{Szegedy14} first found that, with elaborate strategies, deep neural networks can be easily fooled by test images with imperceptible noise~\cite{Szegedy14}, named adversarial examples. These images could be further classified as image-dependent adversarial examples and image-agnostic adversarial examples (universal adversarial examples).

\subsubsection{Image-dependent adversarial examples}
Compared with the initial adversarial examples generated by a box-constrained LBFGS method~\cite{Szegedy14}, Goodfellow \emph{et al}.\ \cite{goodfellow2014explaining} proposed a more time-saving method, FGSM, which builds an adversarial example by performing one-step gradient updating along the direction of the sign of the gradient at each pixel. Kurakin \emph{et al}.\ \cite{Kurakin17} further developed FGSM to generate iterative attacks, and achieved a higher attack success rate than FGSM in a white-box setting. Similarly, another efficient iterative attack method proposed by Moosavi-Dezfooli \emph{et al}.\ \cite{moosavi2016deepfool}, called DeepFool, generated the minimal perturbation in each step by moving towards the lineared decision boundary~\cite{moosavi2016deepfool}. In addition to white-box attacks, transferable black-box attacks~\cite{liu2016delving, papernot2017practical,dong2018boosting,dong2019evading,xie2019improving,wu2020skip} are more practical in real-world situations, since this type of attack can be applied in a fully black-box manner without any queries on the target system.  

\subsubsection{Universal adversarial examples}
Moosavi-Dezfooli \emph{et al}.\ \cite{moosavi2017universal} proved the existence of image-agnostic adversarial attacks, which can create a universal adversarial perturbation (UAP) for all the images of the database. Hayes \emph{et al}.\ \cite{hayes2018learning} investigated the capacity for generative models to craft UAPs and demonstrate that their attack method improves previous methods~\cite{moosavi2017universal} in terms of crafting more transferable UAPs. Mopuri \emph{et al}.\ \cite{reddy2018nag} proposed a generative method aiming to model the distribution of adversarial perturbations by generating a wide variety of UAPs. In concurrent work, Poursaeed \emph{et al}.\ \cite{poursaeed2018generative} proposed a unifying framework based on generative models for creating universal and image-dependent perturbations and improve the state-of-the-art performance of UAPs. Mopuri \emph{et al}.\ \cite{mopuri2017fast,mopuri2018generalizable} first proposed a data-free objective, named GD-UAP, to learn perturbations that can adulterate the deep features extracted by multiple layers. GD-UAP, especially GD-UAP with data priors; it demonstrated impressive fooling rates and surprising transferability across various deep models with different architectures, regularizers and underlying tasks. 
 


At present, the most relevant works to ours are the  class-wise/discriminative methods~\cite{gupta2019method,zhang2020cd}. Gupta \emph{et al}.\ \cite{gupta2019method} proposed a type of class-wise UAP, using the linearity of the decision boundaries of deep neural networks. Using the absolute accuracy drop as an
evaluation metric, Zhang \emph{et al}.\ \cite{zhang2020cd} proposed a class-discriminative universal attack to generate a perturbation that can fool a target network to misclassify only the target classes, while having limited influence on the remaining classes. Note that a class-discriminative universal attack is not suitable for the privacy protection task since it actually sacrifices some protection effects on the target classes in exchange for inhibition effects on the remaining classes. Previous class-wise/discriminative methods~\cite{gupta2019method,zhang2020cd} are inspiring; however, due to a heavy reliance on class information, they cannot be applied to the open-set face recognition task directly. Even if face recognition models suffer from a lack of class information, the proposed OPOM method models the feature subspace of an identity with a data prior and therefore can generate class-wise universal adversarial perturbations by optimizing the deep features in the direction away from the feature subspace of the source identity, which is exactly the advantage of OPOM.

\begin{table}[t]
	\renewcommand\arraystretch{1.0}
	\center
	\caption{Comparison of methods to generate privacy masks with adversarial examples to protect images against malicious face recognition systems.}
	\label{table:table_ref}
	\begin{tabular}{@{}lcccc@{}}
		\toprule
		Methods    & Inference & Open-set & Blackbox & \begin{tabular}[c]{@{}c@{}}Person-specific\\ (Class-wise)\end{tabular} \\ \midrule
		Oh \emph{et al}.\ \cite{joon2017adversarial}         & \textbf{Yes}                                                          & No       & No       & No              \\ \midrule
		P-FGVM~\cite{chatzikyriakidis2019adversarial}      & \textbf{Yes}                                                          & No       & No       & No              \\ \midrule
		Fawkes~\cite{shan2020fawkes}     & No                                                           & No       & \textbf{Yes}      & No              \\ \midrule
		TIP-IM~\cite{yang2020towards}     & \textbf{Yes}                                                          & \textbf{Yes}      & \textbf{Yes}      & No              \\ \midrule
		APF~\cite{zhang2020adversarial}         & \textbf{Yes}                                                          & \textbf{Yes}      & \textbf{Yes}      & No              \\ \midrule
		LowKey~\cite{cherepanova2020lowkey}     & \textbf{Yes}                                                          & \textbf{Yes}      & \textbf{Yes}      & No              \\ \midrule
		This paper & \textbf{Yes}                                                        & \textbf{Yes}      & \textbf{Yes}      & \textbf{Yes}             \\ \bottomrule
	\end{tabular}
\end{table}

\subsection{Adversarial Privacy Protection}
Inspired by the adversarial examples, which can mislead deep models to generate incorrect outputs while maintaining the visual quality of images, some works consider generating adversarial perturbations of face images for privacy protection~\cite{joon2017adversarial,chatzikyriakidis2019adversarial,shan2020fawkes,yang2020towards,zhang2020adversarial,cherepanova2020lowkey}. 

Oh \emph{et al}.\ \cite{joon2017adversarial} made an initial attempt to use adversarial perturbations for privacy protection from a game theory perspective. For better visualization effects, Chatzikyriakidis \emph{et al}.\ \cite{chatzikyriakidis2019adversarial} proposed P-FGVM to generate white-box privacy masks by considering minimal facial image distortion based on iterative attacks~\cite{Kurakin17}. Fawkes~\cite{shan2020fawkes} incorporated adversarial perturbations in the training process by poisoning some target training samples; and therefore misled the model on the poisoned targets in the inference period. Fawkes reported good performance on commercial APIs, which drew much attention in the research community. However, compared with the poisoning in the training process method, it is more applicable to design privacy masks that can be directly used~\cite{yang2020towards,zhang2020adversarial,cherepanova2020lowkey} in terms of privacy protection for regular users. For a higher privacy protection success rate, Yang \emph{et al}.\ \cite{yang2020towards} designed adversarial privacy masks that were directly used on the probe images, with a targeted optimization objective, named TIP-IM, under the assumption of unknown gallery images. To prevent privacy leakage when generating privacy masks, Zhang \emph{et al}.\ \cite{zhang2020adversarial} proposed an end-cloud collaborated adversarial attack solution, named APF, where the original face images were only available at the user end. In addition, Valeriia \emph{et al}.\ \cite{cherepanova2020lowkey} applied adversarial privacy masks on the gallery faces so that the probe images could not be recognized correctly. 

A comparison of the aforementioned methods~\cite{joon2017adversarial,chatzikyriakidis2019adversarial,shan2020fawkes,yang2020towards,zhang2020adversarial,cherepanova2020lowkey} and our methods is listed in Table~\ref{table:table_ref}, in terms of whether the method can be directly used in the inference process, whether the method is applied in an open-set face recognition setting, whether the privacy mask can confront black-box models, and whether the privacy mask is person-specific (class-wise) universal for convenient usage. Despite the clear motivation and good performance~\cite{joon2017adversarial,chatzikyriakidis2019adversarial,shan2020fawkes,yang2020towards,zhang2020adversarial,cherepanova2020lowkey}, to the best of our
knowledge, no works before have considered the person-specific privacy protection for high-efficiency, which we believe could revolutionize the future for face privacy protection. 

\section{Person-Specific Privacy Masks}
\label{sec:method}

\subsection{Problem Formulation}
The objective of a person-specific adversarial mask is to craft a perturbation $\Delta {X}$ that can fool a variety of deep face recognition models $f(\cdot)$ for all the face images $ {{X}^k} =\left\{ X_1^k,X_2^k,\ldots ,X_i^k, \ldots \right\}$ of identity $k$, so that any face image $X_{i}^{k}$ can conceal the identity ${{X}^k}$ from deep face recognition models $f(\cdot)$ with this privacy mask $\Delta {X}$. That is, we find a $\Delta {X}$ so that $X_{i}^{k} \in {{X}^k}$, 
\begin{equation}
\label{equ:obj}
\begin{gathered}
D( f( X_{i}^{k}+\Delta {X}), f_{{{X}^k}}) > t, \,\, \lVert \Delta {X} \rVert _{\infty}<\varepsilon,
\end{gathered}
\end{equation}where $f(X_{i}^{k})\in \mathbb{R}^d$ is the normalized feature of image $X_{i}^{k}$, $f_{{{X}^k}}$ denotes the feature subspace of identity $k$, and $t$ is the distance threshold to decide whether a pair of face images belongs to the same identity. $D(x_1,x_2)$ denotes the distance between $x_1$ and $x_2$, that is, the shortest distance between point $x_1$ and subspace $x_2$. When $x_2$ denotes a point, we use $D(x_1,x_2)$ to represent the normalized Euclidean distance or cosine distance commonly used in face recognition. $\varepsilon$ limits the maximum deviation of the privacy mask. 

\subsection{One person one mask (OPOM)}
In the above problem formulation as Equation~\ref{equ:obj}, the key point is the formulation of $f_{{{X}^k}}$. Note that for open-set face recognition models that are more like feature extractors, it is likely that the persons to be protected do not belong to the training databases. That is, we cannot obtain identity information from the face recognition models directly as previous works~\cite{gupta2019method,zhang2020cd} did in some close-set tasks where the class information of the model can be used to generate class-wise universal perturbations. 

Therefore, the only choice is to describe an identity $f_{{{X}^k}}$ from the provided face image set ${{X}^k}$. For a specific identity, when more face images are provided, the image set ${{X}^k}$ will be more complete, and the feature subspace $f_{{{X}^k}}$ will be more precise. However, incorporating more images for generating adversarial masks is completely at odds with our expectations, since we aim to use as few face images as we can to generate a person-specific adversarial mask for privacy protection. Considering the above function, we explore some approximation methods for $f_{{{X}^k}}$, with a limited number ${n_k}$ of face images $\tilde{{{X}^k}} = \left\{ X_1^k,X_2^k,\ldots ,X_{n_k}^k\right\}$ in practice.

The basic intuition is as follows. With more precise approximation of the identity subspace, the individual universality can be enhanced to some extent. In addition, with more diverse gradient information (\emph{i.e}., gradients of cross-comparisons, images $X_i^K$ and $X_j^K$) incorporated into the approximation equations, the model transferability can be increased to some degree.

To confirm this intuition, we conduct an analytical experiment as shown in Figure~\ref{fig:intro_toy}. Specifically, we describe individual universality using the protection success rate of the unfamiliar images on the local surrogate model, that is, evaluate the masks with the same model and different images. Model transferability can be described as the average protection success rate of the provided training dataset on different black-box models, that is, evaluating the privacy masks with the same images and different models. 

We empirically find that, compared with single points representing the identity (``FI-UAP'', represented with the blue marker), with more precise approximation (represented with the red markers), both individual universality and model transferability can be improved. The empirical experimental results are consistent with the intuition, that is, a better approximation method will lead to more effective privacy protection. Next, we introduce the approximation methods in details. 

\subsubsection{Approximation methods of the feature subspace}

\textbf{Affine Hulls.} The affine hull implicitly augments existing face images, by treating any affine combination of normalized deep features as a valid feature for the identity, so that limited face images can realize their full potential. The affine hull of deep features for a specific identity can be expressed as
\begin{equation}
\label{equa:H_equ}
H(f_{\tilde{{{X}^k}}}) =\left\{ x=\sum_{i=1}^{n_k}{\alpha _{i}^{k}f( X_{i}^{k})} {\bigg|} \sum_{i=1}^{n_k}{\alpha _{i}^{k}=1} \right\}, 
\end{equation}where $f( X_{i}^{k})$ and $\alpha _{i}^{k}$ ($i=1,\ldots ,n_k$) are the features and coefficients to describe the identity $k$.

In this way, we can generate the privacy mask $\Delta {X}$ with limited face images for identity $k$ by optimizing the following objective,
\begin{equation}
\label{equa:loss}
\begin{gathered}
\Delta {X} = \mathop {\arg \max }\limits_{\Delta {X}} \sum_{i=1}^{n_k}{D( f( X_{i}^{k}+\Delta {X}), H(f_{\tilde{{{X}^k}}}) ) )},\,\,
\lVert \Delta {X} \rVert _{\infty}<\varepsilon,
\end{gathered}
\end{equation}where $n_k$ denotes the number of face images, $H(f_{\tilde{{{X}^k}}})$ is the affine hull of the normalized features as Equation~\ref{equa:H_equ}. 

To calculate the distance $D( f( X_{i}^{k}+\Delta {X}), H(f_{\tilde{{{X}^k}}}) )$ in Equation~\ref{equa:loss}, we can rewrite $H(f_{\tilde{{{X}^k}}})$ as follows to parametrize the affine hull, 
\begin{equation}
\label{equa:H_equ_svd}
H(f_{\tilde{{{X}^k}}}) \,\,=\left\{ x=U^kV^k+\mu ^k {\bigg|} V^k\in \mathbb{R}^{n_k} \right\}, 
\end{equation}where $\mu^k = \frac{1}{n_k}\sum_{i=1}^{n_k}{f( X_{i}^{k})}$, $U^k \in \mathbb{R}^{d \times n_k} $ is the orthonormal basis for the directions spanned by the affine hull, obtained by using singular value decomposition (SVD) to $\left[ f( X_{1}^{k}) -\mu ^k,\ldots ,f(X_{n_k}^{k})-\mu ^k \right]$, and $V^k \in \mathbb{R}^{n_k}$ is a vector of free parameters that provides coordinates for the orthonormal basis. With Equation~\ref{equa:H_equ_svd}, we can calculate the distance $D( f( X_{i}^{k}+\Delta {X}), H(f_{\tilde{{{X}^k}}}) )$ in Equation~\ref{equa:loss} as
\begin{equation}
\min_{V^k} \lVert U^kV^k+\mu ^k-f( X_{i}^{k}+\Delta {X}) \rVert, 
\end{equation} which can be written as a standard least squares problem
\begin{equation}\label{equ:affinehull_leaset_squares}\min_{V^k} \lVert U^kV^k-( f( X_{i}^{k}+\Delta {X}) -\mu ^k) \rVert,\end{equation}and the solution is \begin{equation}\label{equ:vk_solution}V^k = ( {U^k}^T{U^k}) ^{-1}{U^k}^T( f( X_{i}^{k}+\Delta {X} ) -\mu ^k).\end{equation} Finally, we can generate the privacy mask $\Delta {X}$ by transforming Equation~\ref{equa:loss} as
\begin{equation}
\label{equ:afinehull_final}
\begin{gathered}
\Delta {X} = \mathop {\arg \max }\limits_{\Delta {X}} J_{A\!H}\hfill \\
\qquad\quad\!\!\!\!\!\!\!\!=\mathop {\arg \max }\limits_{\Delta {X}} \sum_{i=1}^{n_k}{\lVert U^k{V_i^k}-( f( X_{i}^{k}+\Delta {X}) -\mu ^k) \rVert},\,\,
\lVert \Delta {X} \rVert _{\infty}<\varepsilon,
\end{gathered}\end{equation}where $U^k$ and $\mu ^k$ can be calculated as mentioned above, and $V_i^k$ is the solution of Equation~\ref{equ:affinehull_leaset_squares}, with the value shown in Equation~\ref{equ:vk_solution}.

\textbf{Class Centers and Convex Hulls.} 
Although affine hulls can provide an implicit augmentation for provided face images, this approximation may be too loose since some points in the hull may not be valid and may cause opposite effects when generating the mask. Therefore, we introduce lower and upper bounds $L$ and $U$ on the allowable $\alpha_i^k$ coefficients in
Equation~\ref{equ:obj} to control the looseness,
\begin{equation}
\label{equa:RH_equ}
H(f_{\tilde{{{X}^k}}}) = \left\{ x=\sum_{i=1}^{n_k}{\alpha _{i}^{k}f( X_{i}^{k})} {\bigg|} \sum_{i=1}^{n_k}{\alpha _{i}^{k}=1}, L\leqslant \alpha _{i}^{k}\leqslant U \right\},
\end{equation}where $f( X_{i}^{k})$ and $\alpha _{i}^{k}$ ($i=1,\ldots ,n_k$) are the features and coefficients to describe the identity $k$. 

The approximation will be the affine hull if $L = - \infty$ and $U = \infty$. Otherwise, the incorporation of $L$ and $U$ can reduce the over-large region. Another interesting point is $L=U= 1/{n_k}$. In this case, the identity is exactly described as the mean feature
\begin{equation}H(f_{\tilde{{{X}^k}}}) = \frac{1}{n_k}\sum_{i=1}^{n_k}{f( X_{i}^{k})},\end{equation}which is a popular method~\cite{wen2016discriminative} to approximate class centers. In this way, the privacy mask $\Delta {X}$ can be generated by 
\begin{equation}
\label{equ:center_final}
\begin{gathered}
\Delta {X} = \mathop {\arg \max }\limits_{\Delta {X}} J_{C\!C}\hfill \\
\qquad\quad\!\!\!\!\!\!\!\!=\mathop {\arg \max }\limits_{\Delta {X}} \sum_{i=1}^{n_k}{\lVert \frac{1}{n_k}\sum_{j=1}^{n_k}{f( X_{j}^{k})} - f( {X_{i}^{k}+\Delta {X}} ) \rVert},\,\,
\lVert \Delta {X} \rVert _{\infty}<\varepsilon.
\end{gathered}\end{equation}   

Note that if $L=0$ and $U=1$, then $H(f_{\tilde{{{X}^k}}})$ actually approximates the identity with the convex hull (the smallest convex set) of features $f( X_{i}^{k})$, which is also the most effective approximation method we will empirically demonstrate. For the convex hull, \emph{i.e}., with Equation~\ref{equa:RH_equ} and $L=0$, $U=1$, we can calculate $D( f( X_{i}^{k}+\Delta {X}), H(f_{\tilde{{{X}^k}}}) )$ in Equation~\ref{equa:loss} as a least squares problem with box constraints,
\begin{equation}
\label{equ:sovconv}
\min_{{\alpha^k}} \lVert {F^kA^k} - f( {X_{i}^{k}+\Delta {X}} ) \rVert , \!\!\quad s.t.\!\!\quad {A^k} \succeq 0, {1}^T{A^k} =1,
\end{equation}where ${F^k} \in \mathbb{R}^{d \times n_k}$ is a matrix with columns $f({X_{i}^{k}})$, and ${A^k} \in \mathbb{R}^{n_k}$ is a vector containing
the corresponding coefficients ($\alpha _{i}^{k}$ of Equation~\ref{equa:RH_equ}). Then, the privacy mask $\Delta {X}$ can be generated by 
\begin{equation}
\label{equ:convhull_final}
\begin{gathered}
\Delta {X} = \mathop {\arg \max }\limits_{\Delta {X}} J_{CH}\hfill \\
\qquad\quad\!\!\!\!\!\!\!\!=\mathop {\arg \max }\limits_{\Delta {x}} \sum_{i=1}^{n_k}{\lVert {F^kA_i^k} - f( {X_{i}^{k}+\Delta {X}} ) \rVert},\,\,
\lVert \Delta {X} \rVert _{\infty}<\varepsilon,
\end{gathered}\end{equation}where ${A_i^k}$ can be solved by Equation~\ref{equ:sovconv}.

\begin{algorithm}[htbp]
	\caption{OPOM-AffineHull}
	\label{alg:1}
	\LinesNumbered
	\KwIn{Face images $\tilde{{{X}^k}} = \left\{ X_1^k,X_2^k,\ldots ,X_{n_k}^k\right\}$ of identity $k$, deep face model $f(\cdot)$, maximum deviation of perturbations $\varepsilon$, maximum iterative steps $N_{max}$.}	
	\textbf{Initialize:} $\Delta X_0\sim U( -\varepsilon ,\varepsilon ) $, $N = 0$, $g_N=0$\;
	\While{step $N < N_{max}$}{
		Parametrize the affine hull with $U^k$ and $\mu^k$\; 
		
		Calculate $V_i^k$ by Equation~\ref{equ:vk_solution}\;
		
		
		$J_{A\!H}\!=\!\sum_{i=1}^{n_k}{\!\lVert U^k{V_i^k}-( f( X_{i}^{k}+\Delta {X_{N}}) \!-\mu ^k) \rVert}$(Equation~\ref{equ:afinehull_final})\;
	
		$g_{N+1}=\nabla_{X_i^k+\Delta X_N}  J_{A\!H}$\;
		
		$\Delta X_{N+1} = C_{\varepsilon}(\Delta X_{N} + sign(g_{N+1}))$, $N=N+1$\;
	}
	\KwOut{Privacy mask $\Delta X_{N_{max}}$ for identity $k$.}	
\end{algorithm}

\begin{algorithm}[htbp]
	\caption{OPOM-ConvexHull}
	\label{alg:2}
	\LinesNumbered
	\KwIn{Face images $\tilde{{{X}^k}} = \left\{ X_1^k,X_2^k,\ldots ,X_{n_k}^k\right\}$ of identity $k$, deep face model $f(\cdot)$, maximum deviation of perturbations $\varepsilon$, maximum iterative steps $N_{max}$.}	
	\textbf{Initialize:} $\Delta X_0\sim U( -\varepsilon ,\varepsilon ) $, $N = 0$, $g_N=0$\;
	\While{step $N < N_{max}$}{
		Calculate $A_i^k$ by solving Equation~\ref{equ:sovconv}\;
		
		$J_{C\!H}\!=\!\sum_{i=1}^{n_k}{\!\lVert {F^kA_i^k} - f( {X_{i}^{k}+\Delta {X_N}} ) \rVert}$ (Equation~\ref{equ:convhull_final})\;
		
		$g_{N+1}=\nabla_{X_i^k+\Delta X_N} J_{C\!H}$\;
		
		$\Delta X_{N+1} = C_{\varepsilon}(\Delta X_{N} + sign(g_{N+1}))$, $N=N+1$\;	
	}
	\KwOut{Privacy mask $\Delta X_{N_{max}}$ for identity $k$.}	
\end{algorithm}

\subsubsection{Generation of Privacy Masks}
\label{sec:trans}
With the approximation methods for modeling the feature subspace of an identity using affine hulls, class centers and convex hulls, we formulate the person-specific privacy mask generation to optimization problems as Equation~\ref{equ:afinehull_final}, Equation~\ref{equ:center_final} and Equation~\ref{equ:convhull_final}, named OPOM-AffineHull, OPOM-ClassCenter and OPOM-ConvexHull. The optimization problems can be solved by an iteratively signed gradient ascent as in previous works~\cite{zhong2020towards}, which is known to be effective for generating adversarial attacks towards deep models~\cite{madry2017towards}. 

To further detail the generation of person-specific masks, we conclude the proposed OPOM-AffineHull and OPOM-ConvexHull in Algorithm~\ref{alg:1} and Algorithm~\ref{alg:2}, where $C{_{\varepsilon }}(X) = \min ( \varepsilon ,\max (- \varepsilon ,X))$. For brevity, we omit OPOM-ClassCenter. To calculate the distance between the protected features and the original feature subspace in each iterative step, the coordinates $V_i^k$ or coefficients $A_i^k$ should be calculated at the beginning of each iterative step. We provide a closed-form solution for $V_i^k$ in Equation~\ref{equ:vk_solution}. For $A_i^k$, although there is no closed-form solution, it can be efficiently solved with the convex optimization toolbox CVX~\cite{diamond2016cvxpy,agrawal2018rewriting}. 

\subsubsection{Combination with Model Transferability Methods}
To generate more transferable person-specific privacy masks from a single source model to protect face images against black-box models, model transferability methods, such as the momentum boosting method~\cite{dong2018boosting} and DFANet~\cite{zhong2020towards}, can be incorporated into OPOM method. 

The momentum boosting method~\cite{dong2018boosting} integrates the momentum term into the attack process to stabilize the update directions and escape from poor local maxima. Specifically, $g_{N+1}$ in Algorithm~\ref{alg:1} and Algorithm~\ref{alg:2} gathers gradients of the first $N$ iterations with a decay factor $\mu$, \begin{equation}
g_{N+1}=\mu \cdot g_N+\frac{\nabla _{X_{i}^{k}+\Delta X_N}J}{\lVert \nabla _{X_{i}^{k}+\Delta X_N}J \rVert},
\end{equation}
where $\mu$ is usually set to 1 following the original paper. 

DFANet~\cite{zhong2020towards} converts the surrogate model $f(\cdot)$ in the N-th step of the adversarial example generation to different models $\tilde{f}^N(x)$ of the $N$-th step by incorporating dropout layers. Specifically, for a face recognition model $f(\cdot)$ composed of convolutional layers, given the output $o_i\in \mathbb{R}^n$ from the $i$-th convolutional layer, a mask $M_i\in \mathbb{R}^n$ is generated with each element $m_i$ independently sampled from a Bernoulli distribution with probability $p_d$:\begin{equation}\label{drop}m_i\sim Bernoulli\left( p_d \right), \qquad m_i\in M_i.\end{equation}Then, this mask is applied to modify the output as $o_i = M_i \times o_i$, where $\times$ denotes the Hadamard product. 

\subsection{Comparison methods}
\label{sec:compare}
In addition to OPOM, in this paper we explore other methods that can generate universal adversarial perturbations or class-universal adversarial perturbations, for the potential capabilities in the person-specific privacy protection task.

\subsubsection{GD-UAP} To generate universal perturbations, GD-UAP seeks $\Delta {X}$ to produce maximal spurious activations at each layer of a model. Although GD-UAP is a data-free optimization for crafting image-agnostic perturbations, according to the original paper~\cite{mopuri2018generalizable}, for the best performance of GD-UAP, it can be utilized with simple data priors, such as target data samples. Therefore, for a fair comparison with OPOM, $\tilde{{{X}^k}} = \left\{ X_1^k,X_2^k,\ldots ,X_{n_k}^k\right\}$ is used as the data priors to generate a privacy mask for identity $k$, 
\begin{equation}
\begin{gathered}
\Delta {X} = \mathop {\arg \min }\limits_{\Delta {X}} J_{G\!D-U\!\!A\!P}\hfill \\
\!\!\!\!\!\!\!\!\qquad\quad=\mathop {\arg \min }\limits_{\Delta {X}} -\sum_i^{n_k}{\log \left( \prod_{j=1}^K{\lVert l_j ( X_{i}^{k}+\Delta {X} ) \rVert} \right)}, \lVert \Delta {X} \rVert _{\infty}<\varepsilon,
\end{gathered}\end{equation}where $l_j(X_{i}^{k}+\Delta {X})$ is the activation in the output tensor (after the non-linearity) at layer $j$ when $X_{i}^{k}+\Delta {X}$ is fed to the network $f(\cdot)$, and $K$ is the number of layers.  

\subsubsection{FI-UAP, FI-UAP+ and FI-UAP-all} One straightforward idea to generate class-wise universal adversarial perturbations towards open-set face recognition models is actually a similar way in the spirit of UAP~\cite{moosavi2017universal}, which leverages a set of images to seek the minimal perturbation iteratively by DeepFool~\cite{moosavi2016deepfool} and aggregates them to the universal perturbations. The feature iterative attack method (FIM)~\cite{zhong2019adversarial,zhong2020towards}, which has been demonstrated to be an applicable method for face recognition models, can also be used in this way. Specifically, we aggregate gradients of $\tilde{{{X}^k}} = \left\{ X_1^k,X_2^k,\ldots ,X_{n_k}^k\right\}$ to generate person-specific masks, 
\begin{equation}
\label{equ:fiuap_final}
\begin{gathered}
\Delta {X} = \mathop {\arg \max }\limits_{\Delta {X}} J_{F\!I-U\!\!A\!P}\hfill \\
\qquad\quad\!\!\!\!\!\!\!\!=\mathop {\arg \max }\limits_{\Delta {X}} \sum_{i=1}^{n_k}{\lVert f( X_{i}^{k}) - f({X_{i}^{k}+\Delta {X}})   \rVert},\,\,
\lVert \Delta {X} \rVert _{\infty}<\varepsilon,
\end{gathered}\end{equation}and name this method FI-UAP, which can be considered one approach for using the single point to model the identity information.

FI-UAP can be enhanced by incorporating intra-class interactions, which we refer to as FI-UAP+. 

\begin{equation}
	\label{equ:fiuap+_final}
	\begin{gathered}
		\Delta {X} = \mathop {\arg \max }\limits_{\Delta {X}} J_{F\!I-U\!\!A\!P+}\hfill \\
		\qquad\quad\!\!\!\!\!\!\!\!=\mathop {\mathrm{arg}\max} \limits_{\Delta X}\sum_{i=1}^{n_k}{\sum_{j=1}^{n_k}{\left\| f(X_{i}^{k})-f(X_{j}^{k}+\Delta X) \right\|}}
		,\,\,
		\lVert \Delta {X} \rVert _{\infty}<\varepsilon.
	\end{gathered}\end{equation}

Furthermore, FI-UAP can be extended to generate a universal mask if all the training images are used, which we refer to as FI-UAP-all.  

\subsubsection{GAP}
Generative adversarial perturbations (GAP)~\cite{poursaeed2018generative}  pass a fixed pattern sampled from a uniform distribution through the generator to generate universal adversarial perturbations, which can be added to a set of images to mislead models. We strictly follow the original paper using the ResNet Generator but replace the label-level adversarial losses as feature-level ones, as Equation~\ref{equ:fiuap_final}, to make it more suitable for face recognition models~\cite{zhong2020towards}. 

\section{Experiments}
\label{sec:experiment}
In this section, we first introduce the experimental settings. Next, we report the protection performance of the comparison methods and the proposed OPOM method, explore the performance of OPOM combined with the transferability methods, and show the protection effects against commercial APIs. Finally, we discuss the proposed method in the following aspects: analyses of the failure cases, the strengths and weaknesses of the person-specific (class-wise) universal masks compared with image-specific masks and universal masks, the practicality of OPOM in the privacy protection of a video dataset, and the diversity of privacy masks for the potential leakage of the used privacy masks.

\begin{table*}[]
	\center
	\renewcommand\arraystretch{1.0}
	\caption{Comparison of different methods to generate person-specific privacy masks ($\varepsilon = 8$) from a single source model to protect face images against black-box models. We report Top-1 and Top-5 protection success rate (\%) under 1:N identification setting of the Privacy-Commons dataset. The higher protection success rate is better.}
	\label{table:table_common}
	\scalebox{0.99}{
	\begin{tabular}{@{}c|c|cccccccccccc@{}}
		\toprule
		\multirow{3}{*}{\textbf{Source}} & \multirow{3}{*}{\textbf{Method}} & \multicolumn{12}{c}{\textbf{Target }}                                                                                                                                                                                                                                                                                                          \\
		&                                  & \multicolumn{2}{c|}{\textbf{ArcFace}}                    & \multicolumn{2}{c|}{\textbf{CosFace}}                    & \multicolumn{2}{c|}{\textbf{SFace}}                      & \multicolumn{2}{c|}{\textbf{MobileNet}}                  & \multicolumn{2}{c|}{\textbf{SENet}}                      & \multicolumn{2}{c}{\textbf{Inception-ResNet}} \\
		&                                  & \textbf{Top-1}   & \multicolumn{1}{c|}{\textbf{Top-5}}   & \textbf{Top-1}   & \multicolumn{1}{c|}{\textbf{Top-5}}   & \textbf{Top-1}   & \multicolumn{1}{c|}{\textbf{Top-5}}   & \textbf{Top-1}   & \multicolumn{1}{c|}{\textbf{Top-5}}   & \textbf{Top-1}   & \multicolumn{1}{c|}{\textbf{Top-5}}   & \textbf{Top-1}        & \textbf{Top-5}        \\ \midrule
		\multirow{8}{*}{Softmax}                
					
		& GD-UAP  & 6.3 & \multicolumn{1}{c|}{2.7}  
		&  3.1 & \multicolumn{1}{c|}{1.6} 
		&  3.0 & \multicolumn{1}{c|}{1.5} 
		&  8.7 & \multicolumn{1}{c|}{3.5}
		&  5.8 & \multicolumn{1}{c|}{2.7}
		&  3.0 & 1.5      \\

		& GAP  & 30.7 & \multicolumn{1}{c|}{20.7}          
		&  19.7 & \multicolumn{1}{c|}{13.9}           
		&  23.9 & \multicolumn{1}{c|}{15.9}           
		&  33.6 & \multicolumn{1}{c|}{20.8} 
		&  31.0 & \multicolumn{1}{c|}{20.1} 
		&  13.1 & 7.4 \\

		& FI-UAP  & 72.3 & \multicolumn{1}{c|}{62.4}          
		& 63.5 & \multicolumn{1}{c|}{53.3} 
		& 70.3 & \multicolumn{1}{c|}{61.9} 
		& 73.9 & \multicolumn{1}{c|}{61.8} 
		& 77.4 & \multicolumn{1}{c|}{67.6}
		& 52.4 & 40.7 \\
				
		& FI-UAP+  & 76.9 & \multicolumn{1}{c|}{67.8}          
		& 69.2 & \multicolumn{1}{c|}{60.3}
		& 75.1 & \multicolumn{1}{c|}{67.3}           
		& 78.3 & \multicolumn{1}{c|}{67.8} 
		& 82.1 & \multicolumn{1}{c|}{73.3} 
		& 57.2 & 45.6 \\
		
		& FI-UAP-all  & 53.4 & \multicolumn{1}{c|}{43.2} 
		& 38.5 & \multicolumn{1}{c|}{30.3} 
		& 39.8 & \multicolumn{1}{c|}{31.0} 
		& 54.0 & \multicolumn{1}{c|}{39.4} 
		& 51.3 & \multicolumn{1}{c|}{38.9} 
		& 24.2 & 17.1 \\
		
		& OPOM-AffineHull  & 73.0 & \multicolumn{1}{c|}{63.1} 
		& 63.3 & \multicolumn{1}{c|}{53.8} 
		& 71.1 & \multicolumn{1}{c|}{62.0} 
		& 74.7 & \multicolumn{1}{c|}{63.0} 
		& 77.8 & \multicolumn{1}{c|}{68.6} 
		& 52.7 & 40.8 \\
		
		& OPOM-ClassCenter  & 76.9 & \multicolumn{1}{c|}{67.8}          
		& 69.2 & \multicolumn{1}{c|}{60.3} 
		& 75.0 & \multicolumn{1}{c|}{67.3}           
		& 78.4 & \multicolumn{1}{c|}{67.9}          
		& 82.1 & \multicolumn{1}{c|}{73.4}          
		& 57.1 & 45.6\\
		
		& OPOM-ConvexHull  
		& \textbf{78.0}  & \multicolumn{1}{c|}{\textbf{69.4}}  
		& \textbf{70.2}  & \multicolumn{1}{c|}{\textbf{61.4} }           
		& \textbf{76.1}  & \multicolumn{1}{c|}{\textbf{68.7}}            
		& \textbf{79.2}  & \multicolumn{1}{c|}{\textbf{69.1}}           
		& \textbf{82.9}  & \multicolumn{1}{c|}{\textbf{74.2}}  
		& \textbf{58.7} & \textbf{47.2} \\ \midrule
		
		\multirow{8}{*}{ArcFace}                
		
		& GD-UAP  & 2.6 & \multicolumn{1}{c|}{1.0} 
		& 1.1 & \multicolumn{1}{c|}{0.6}           
		& 1.1 & \multicolumn{1}{c|}{0.4}           
		& 2.9 & \multicolumn{1}{c|}{1.0}          
		& 1.9 & \multicolumn{1}{c|}{0.8}          
		& 1.5 & 0.7
		      \\
		
		& GAP  & 52.7 & \multicolumn{1}{c|}{36.7}          
		& 36.7 & \multicolumn{1}{c|}{26.1}           
		& 41.8 & \multicolumn{1}{c|}{30.6}           
		& 51.8 & \multicolumn{1}{c|}{36.6}          
		& 49.5 & \multicolumn{1}{c|}{36.1}          
		& 28.1 & 18.9 \\
		
		& FI-UAP  & 82.3 & \multicolumn{1}{c|}{75.0}          
		& 71.2 & \multicolumn{1}{c|}{63.6} 
		& 77.3 & \multicolumn{1}{c|}{70.0}           
		& 65.2 & \multicolumn{1}{c|}{50.4}
		& 73.9 & \multicolumn{1}{c|}{63.1}          
		& 56.6 & 45.3 \\
		
		& FI-UAP+  & 85.9 & \multicolumn{1}{c|}{79.6}          
		& 76.2 & \multicolumn{1}{c|}{69.5}  
		& 82.0 & \multicolumn{1}{c|}{75.7} 
		& 69.8 & \multicolumn{1}{c|}{56.7}  
		& 78.7 & \multicolumn{1}{c|}{69.8}          
		& 62.0 & 50.9 \\
		
		& FI-UAP-all  & 60.5 & \multicolumn{1}{c|}{50.9}          
		& 45.1 & \multicolumn{1}{c|}{35.6} 
		& 49.1 & \multicolumn{1}{c|}{39.8}           
		& 46.0 & \multicolumn{1}{c|}{33.1}          
		& 48.4 & \multicolumn{1}{c|}{36.1}
		& 26.6 & 19.0 \\
		
		& OPOM-AffineHull  & 82.9 & \multicolumn{1}{c|}{75.8}          
		& 72.3 & \multicolumn{1}{c|}{64.8}           
		& 78.1 & \multicolumn{1}{c|}{70.9}           
		& 66.4 & \multicolumn{1}{c|}{52.2}          
		& 75.4 & \multicolumn{1}{c|}{64.5}          
		& 58.4 & 46.8 \\
		
		& OPOM-ClassCenter  & 86.0 & \multicolumn{1}{c|}{79.5}          
		& 76.1 & \multicolumn{1}{c|}{69.6}           
		& 82.0 & \multicolumn{1}{c|}{75.8}           
		& 69.6 & \multicolumn{1}{c|}{56.7}          
		& 78.6 & \multicolumn{1}{c|}{69.7}          
		& 61.8 & 51.0 \\
		
		& OPOM-ConvexHull  
		& \textbf{86.5} & \multicolumn{1}{c|}{\textbf{80.1}}          
		& \textbf{76.8} & \multicolumn{1}{c|}{\textbf{70.0}}           
		& \textbf{82.7} & \multicolumn{1}{c|}{\textbf{76.5}}           
		& \textbf{70.5} & \multicolumn{1}{c|}{\textbf{57.3}}          
		& \textbf{79.3} & \multicolumn{1}{c|}{\textbf{70.2}} 
		& \textbf{63.2} & \textbf{52.2} \\ \midrule
		
		\multirow{8}{*}{CosFace}                
		
		& GD-UAP  
		& 3.7 & \multicolumn{1}{c|}{1.4}          
		&  1.1 & \multicolumn{1}{c|}{0.4}           
		&  1.3 & \multicolumn{1}{c|}{0.4}           
		&  4.0 & \multicolumn{1}{c|}{1.5}          
		&  3.2 & \multicolumn{1}{c|}{1.0}          
		&  1.8 & 0.6 \\
		
		& GAP  
		& 59.7 & \multicolumn{1}{c|}{46.7}          
		& 40.2 & \multicolumn{1}{c|}{28.4}           
		& 38.8 & \multicolumn{1}{c|}{27.0}           
		& 41.5 & \multicolumn{1}{c|}{28.4}          
		& 47.2 & \multicolumn{1}{c|}{34.0}          
		& 27.5 & 18.0 \\
		
		& FI-UAP 
		& 80.4 & \multicolumn{1}{c|}{72.3}          
		& 72.8 & \multicolumn{1}{c|}{65.1}           
		& 76.7 & \multicolumn{1}{c|}{68.9}          
		& 55.6 & \multicolumn{1}{c|}{41.0}          
		& 66.9 & \multicolumn{1}{c|}{54.6}          
		& 49.9 & 37.1 \\
		
		& FI-UAP+  
		& 85.4 & \multicolumn{1}{c|}{78.5}          
		& 78.6 & \multicolumn{1}{c|}{71.4} 
		& 82.4 & \multicolumn{1}{c|}{75.4}           
		& 61.4 & \multicolumn{1}{c|}{47.6}          
		& 73.0 & \multicolumn{1}{c|}{61.7}          
		& 55.3 & 43.0 \\
		
		& FI-UAP-all  
		& 64.2 & \multicolumn{1}{c|}{51.7}          
		& 48.6 & \multicolumn{1}{c|}{38.4}           
		& 52.1 & \multicolumn{1}{c|}{40.8}           
		& 49.6 & \multicolumn{1}{c|}{36.6}          
		& 56.4 & \multicolumn{1}{c|}{43.3}          
		& 29.3 & 20.5 \\
		
		& OPOM-AffineHull  
		& 82.4 & \multicolumn{1}{c|}{74.5}          
		& 74.3 & \multicolumn{1}{c|}{67.0}           
		& 78.4 & \multicolumn{1}{c|}{71.3}           
		& 58.6 & \multicolumn{1}{c|}{43.8}          
		& 69.0 & \multicolumn{1}{c|}{56.8} 
		& 51.8 & 39.4 \\
		
		& OPOM-ClassCenter  
		& 85.4 & \multicolumn{1}{c|}{78.3}          
		& 78.8 & \multicolumn{1}{c|}{71.6}           
		& 82.4 & \multicolumn{1}{c|}{75.5}           
		& 61.4 & \multicolumn{1}{c|}{47.5}          
		& 73.0 & \multicolumn{1}{c|}{61.5}          
		& 55.2 & 43.0      \\
		
		& OPOM-ConvexHull  
		& \textbf{86.6} & \multicolumn{1}{c|}{\textbf{79.3}}          
		& \textbf{79.5} & \multicolumn{1}{c|}{\textbf{72.7}}           
		& \textbf{83.0} & \multicolumn{1}{c|}{\textbf{76.3}}           
		& \textbf{62.3} & \multicolumn{1}{c|}{\textbf{48.2}}          
		& \textbf{74.0} & \multicolumn{1}{c|}{\textbf{63.1}}         
		& \textbf{56.8} &  \textbf{44.5}     \\ \bottomrule
	\end{tabular}}
\end{table*}

\begin{table*}[]
	\center
	\renewcommand\arraystretch{1.0}
	\caption{Comparison of different methods to generate person-specific privacy masks ($\varepsilon = 8$) from a single source model to protect face images against black-box models. We report Top-1 and Top-5 protection success rate under 1:N identification setting of the Privacy-Celebrities dataset. The higher protection success rate is better.}
	\label{table:table_Celebrities}
	\scalebox{0.99}{
		\begin{tabular}{@{}c|c|cccccccccccc@{}}
		\toprule
		\multirow{3}{*}{\textbf{Source}} & \multirow{3}{*}{\textbf{Method}} & \multicolumn{12}{c}{\textbf{Target }}                                                                                                                                                                                                                                                                                                          \\
		&                                  & \multicolumn{2}{c|}{\textbf{ArcFace}}                    & \multicolumn{2}{c|}{\textbf{CosFace}}                    & \multicolumn{2}{c|}{\textbf{SFace}}                      & \multicolumn{2}{c|}{\textbf{MobileNet}}                  & \multicolumn{2}{c|}{\textbf{SENet}}                      & \multicolumn{2}{c}{\textbf{Inception-ResNet}} \\
		&                                  & \textbf{Top-1}   & \multicolumn{1}{c|}{\textbf{Top-5}}   & \textbf{Top-1}   & \multicolumn{1}{c|}{\textbf{Top-5}}   & \textbf{Top-1}   & \multicolumn{1}{c|}{\textbf{Top-5}}   & \textbf{Top-1}   & \multicolumn{1}{c|}{\textbf{Top-5}}   & \textbf{Top-1}   & \multicolumn{1}{c|}{\textbf{Top-5}}   & \textbf{Top-1}        & \textbf{Top-5}        \\ \midrule
		\multirow{8}{*}{Softmax}                
				
		& GD-UAP  
		& 6.5 & \multicolumn{1}{c|}{2.4}          
		& 3.4 & \multicolumn{1}{c|}{1.2}           
		& 3.6 & \multicolumn{1}{c|}{1.3}           
		& 10.7 & \multicolumn{1}{c|}{3.8}          
		& 5.7  & \multicolumn{1}{c|}{2.1}          
		& 3.1  & 0.9 \\
		
		& GAP  & 41.5 & \multicolumn{1}{c|}{31.3}          
		&  31.7 & \multicolumn{1}{c|}{22.2}           
		&  35.0 & \multicolumn{1}{c|}{25.2}           
		&  53.2 & \multicolumn{1}{c|}{40.7}          
		&  44.5 & \multicolumn{1}{c|}{33.5}          
		&  20.2 & 12.1      \\
		
		& FI-UAP  & 56.5 & \multicolumn{1}{c|}{45.5}          
		& 45.2 & \multicolumn{1}{c|}{34.9}           
		& 52.9 & \multicolumn{1}{c|}{42.4}           
		& 60.2 & \multicolumn{1}{c|}{46.9}          
		& 61.6 & \multicolumn{1}{c|}{49.8}          
		& 37.4 & 26.5 \\
		
		& FI-UAP+  & 62.4 & \multicolumn{1}{c|}{51.6}          
		& 51.8 & \multicolumn{1}{c|}{42.3}           
		& 59.8 & \multicolumn{1}{c|}{49.7}           
		& 66.2 & \multicolumn{1}{c|}{54.7}          
		& 67.8 & \multicolumn{1}{c|}{57.5}          
		& 42.5 & 31.4 \\
		
		& FI-UAP-all  & 47.6 & \multicolumn{1}{c|}{37.9} 
		& 38.4 & \multicolumn{1}{c|}{28.9}           
		& 42.5 & \multicolumn{1}{c|}{33.3}           
		& 59.0 & \multicolumn{1}{c|}{47.8}          
		& 53.9 & \multicolumn{1}{c|}{43.3}          
		& 24.4 & 16.2 \\
		
		& OPOM-AffineHull  & 58.8 & \multicolumn{1}{c|}{47.9}          
		& 47.0 & \multicolumn{1}{c|}{37.2}           
		& 54.8 & \multicolumn{1}{c|}{44.9}           
		& 62.4 & \multicolumn{1}{c|}{49.3}          
		& 63.5 & \multicolumn{1}{c|}{52.1}          
		& 39.0 & 27.8 \\
		
		& OPOM-ClassCenter  & 62.3 & \multicolumn{1}{c|}{51.5} 		        
		& 51.8 & \multicolumn{1}{c|}{42.5}           
		& 59.8 & \multicolumn{1}{c|}{49.8}           
		& 66.2 & \multicolumn{1}{c|}{54.8}          
		& 67.7 & \multicolumn{1}{c|}{57.7}          
		& 42.4 & 31.4 \\
		
		& OPOM-ConvexHull  & \textbf{63.8} & \multicolumn{1}{c|}{\textbf{53.5}}          
		& \textbf{53.6} & \multicolumn{1}{c|}{\textbf{44.0}}           
		& \textbf{61.3} & \multicolumn{1}{c|}{\textbf{51.7}}           
		& \textbf{67.0} & \multicolumn{1}{c|}{\textbf{55.7}}          
		& \textbf{68.9} & \multicolumn{1}{c|}{\textbf{58.9}}          
		& \textbf{44.2} & \textbf{33.2}      \\ \midrule
		
		\multirow{8}{*}{ArcFace}                
		
		& GD-UAP  & 6.5 & \multicolumn{1}{c|}{2.4}          
		& 3.4 & \multicolumn{1}{c|}{1.2}           
		& 4.0 & \multicolumn{1}{c|}{1.3}           
		& 10.0 & \multicolumn{1}{c|}{3.8}          
		& 6.4 & \multicolumn{1}{c|}{2.5}          
		& 3.8 & 1.3 \\
		
		& GAP  & 52.4 & \multicolumn{1}{c|}{41.5}          
		& 40.1 & \multicolumn{1}{c|}{30.2}           
		& 45.4 & \multicolumn{1}{c|}{36.1}           
		& 55.9 & \multicolumn{1}{c|}{43.4}          
		& 50.8 & \multicolumn{1}{c|}{39.5}          
		& 36.5 & 26.3 \\
		
		& FI-UAP  & 62.5 & \multicolumn{1}{c|}{51.5}          
		& 47.9 & \multicolumn{1}{c|}{37.8}           
		& 55.6 & \multicolumn{1}{c|}{45.8}           
		& 50.0 & \multicolumn{1}{c|}{35.6}          
		& 54.3 & \multicolumn{1}{c|}{41.0}          
		& 37.7 & 26.3 \\
		
		& FI-UAP+  & 68.7 & \multicolumn{1}{c|}{59.3}          
		& 55.2 & \multicolumn{1}{c|}{46.1} 
		& 63.0 & \multicolumn{1}{c|}{54.2}           
		& 56.4 & \multicolumn{1}{c|}{42.5}         
		& 60.9 & \multicolumn{1}{c|}{49.0}          
		& 43.6 & 31.8 \\
		
		& FI-UAP-all  & 57.2 & \multicolumn{1}{c|}{46.6}          
		& 43.4 & \multicolumn{1}{c|}{33.4}           
		& 50.6 & \multicolumn{1}{c|}{40.9}           
		& \textbf{58.8} & \multicolumn{1}{c|}{\textbf{45.7}}          
		& 57.3 & \multicolumn{1}{c|}{46.0}          
		& 35.1 & 25.1 \\
		
		& OPOM-AffineHull  & 65.8 & \multicolumn{1}{c|}{55.2}          
		& 51.3 & \multicolumn{1}{c|}{41.3}           
		& 59.1 & \multicolumn{1}{c|}{49.6}           
		& 53.1 & \multicolumn{1}{c|}{39.0}          
		& 57.7 & \multicolumn{1}{c|}{45.1} 
		& 40.5 & 28.4 \\
		
		& OPOM-ClassCenter  & 68.9 & \multicolumn{1}{c|}{59.3}          
		& 55.2 & \multicolumn{1}{c|}{46.1}           
		& 63.0 & \multicolumn{1}{c|}{54.3}           
		& 56.6 & \multicolumn{1}{c|}{42.6}          
		& 61.1 & \multicolumn{1}{c|}{49.0}          
		& 43.8 & 32.0 \\
		
		& OPOM-ConvexHull  & \textbf{69.9} & \multicolumn{1}{c|}{\textbf{60.4}}          
		& \textbf{56.2} & \multicolumn{1}{c|}{\textbf{47.3}}           
		& \textbf{64.1} & \multicolumn{1}{c|}{\textbf{55.1}}           
		& 57.5 & \multicolumn{1}{c|}{43.1}         
		& \textbf{62.3} & \multicolumn{1}{c|}{\textbf{50.0}} 
		& \textbf{44.9} & \textbf{33.2} \\ \midrule
		
		\multirow{8}{*}{CosFace}                
		
		& GD-UAP  & 7.4 & \multicolumn{1}{c|}{3.0}          
		& 3.7 & \multicolumn{1}{c|}{1.3}           
		& 4.2 & \multicolumn{1}{c|}{1.4}           
		& 10.2 & \multicolumn{1}{c|}{4.2}          
		& 6.3 & \multicolumn{1}{c|}{2.4} 
		& 3.8 & 1.3 \\
		
		& GAP  & 53.3 & \multicolumn{1}{c|}{42.7}          
		& 40.4 & \multicolumn{1}{c|}{30.5}           
		& 41.6 & \multicolumn{1}{c|}{31.7}           
		& 48.9 & \multicolumn{1}{c|}{35.9}          
		& 48.1 & \multicolumn{1}{c|}{36.9}          
		& 24.8 & 15.0 \\
		
		& FI-UAP  & 63.7 & \multicolumn{1}{c|}{52.1}          
		& 51.5 & \multicolumn{1}{c|}{41.3}           
		& 57.6 & \multicolumn{1}{c|}{47.2}           
		& 46.6 & \multicolumn{1}{c|}{31.7}          
		& 50.8 & \multicolumn{1}{c|}{37.1}          
		& 33.0 & 21.9 \\
		
		& FI-UAP+  & 68.6 & \multicolumn{1}{c|}{58.6}          
		& 58.4 & \multicolumn{1}{c|}{48.9}           
		& 63.8 & \multicolumn{1}{c|}{54.7} 
		& 51.2 & \multicolumn{1}{c|}{36.4}          
		& 57.2 & \multicolumn{1}{c|}{43.9}          
		& 38.6 & 26.5 \\
		
		& FI-UAP-all  & 58.9 & \multicolumn{1}{c|}{48.2}          
		& 41.4 & \multicolumn{1}{c|}{31.7}           
		& 49.2 & \multicolumn{1}{c|}{39.5} 
		& \textbf{55.7} & \multicolumn{1}{c|}{\textbf{43.5}}          
		& 57.1 & \multicolumn{1}{c|}{\textbf{46.6}}          
		& 33.2 & 23.8 \\
		
		& OPOM-AffineHull  & 66.5 & \multicolumn{1}{c|}{55.6}          
		& 55.4 & \multicolumn{1}{c|}{45.3}           
		& 61.1 & \multicolumn{1}{c|}{51.4}           
		& 49.5 & \multicolumn{1}{c|}{34.7}          
		& 54.7 & \multicolumn{1}{c|}{40.7} 
		& 35.7 & 23.9 \\
		
		& OPOM-ClassCenter  & 68.3 & \multicolumn{1}{c|}{58.4}          
		& 58.5 & \multicolumn{1}{c|}{48.9}           
		& 63.8 & \multicolumn{1}{c|}{54.5}           
		& 51.2 & \multicolumn{1}{c|}{36.3}          
		& 57.0 & \multicolumn{1}{c|}{43.9}          
		& 38.5 & 26.5 \\
		
		& OPOM-ConvexHull  & \textbf{69.9} & \multicolumn{1}{c|}{\textbf{60.6}}         
		& \textbf{59.8} & \multicolumn{1}{c|}{\textbf{50.5}} 
		& \textbf{65.6} & \multicolumn{1}{c|}{\textbf{56.2}}           
		& 53.0 & \multicolumn{1}{c|}{38.1}          
		& \textbf{58.6} & \multicolumn{1}{c|}{45.5}
		& \textbf{40.2} & \textbf{28.0} \\ \bottomrule
	\end{tabular}}
\end{table*}

\begin{figure}[htbp]
	\center
	\includegraphics[width=1\linewidth]{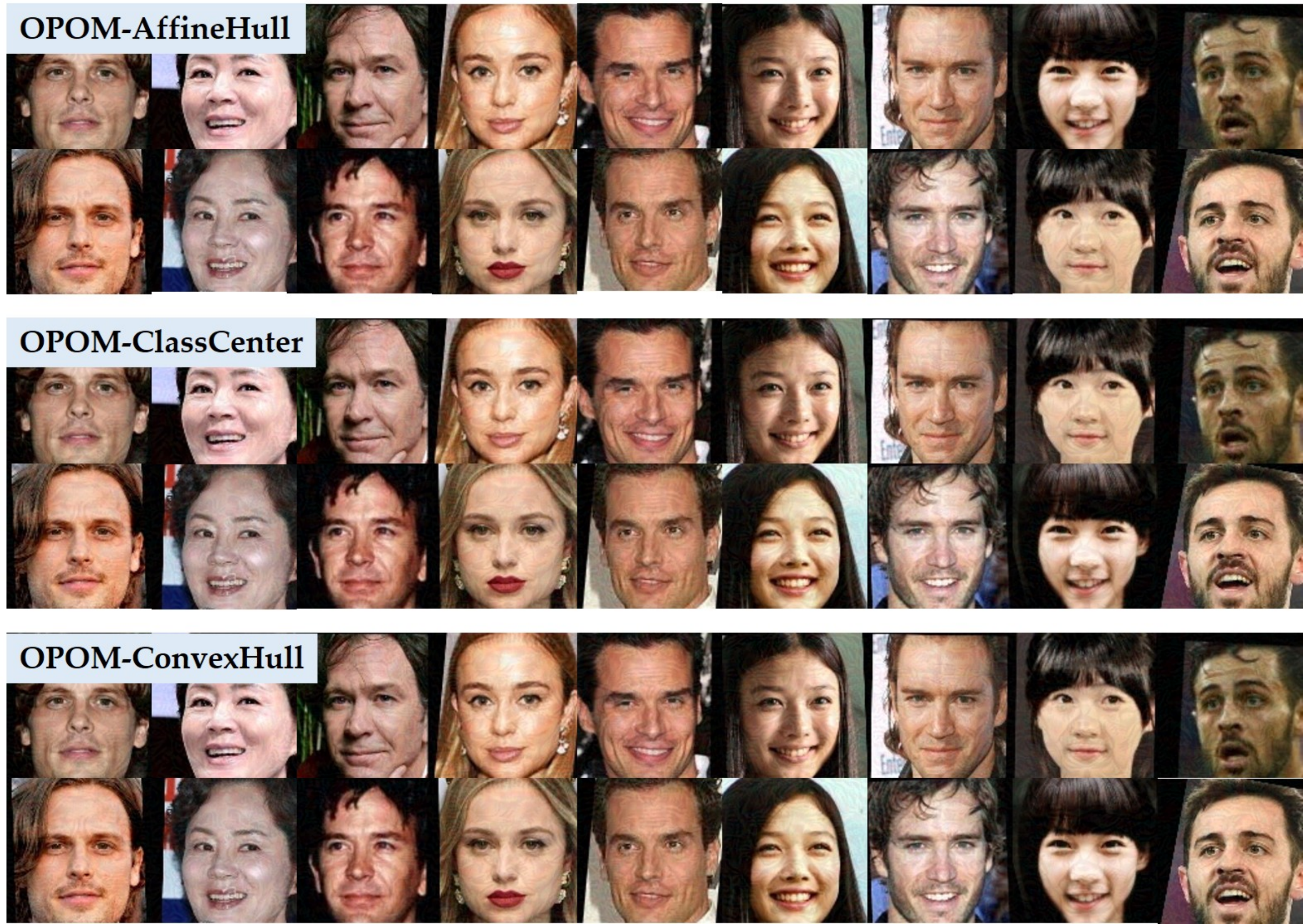}
	\caption{Some protected images in the Privacy-Celebrities dataset with masks ($\varepsilon = 8$), generated by OPOM-AffineHull, OPOM-ClassCenter, and OPOM-ConvexHull respectively.}
	\label{fig:pro}
\end{figure}

\subsection{Experimental Settings}

\subsubsection{Datasets and Evaluation Metrics}
\label{sec:dataset_evaluation} 
In this paper, we simulate a real-word scenario where regular users would like to provide a few images to generate person-specific privacy protection masks. Person-specific masks are generated on the source model with the provided training images. Then, the generated privacy masks can be applied to other test images to protect them against different black-box face recognition models. 

We expect to investigate person-specific adversarial privacy masks, and therefore need identities with relatively enough images to evaluate our method. We build two datasets using common people and celebrities to simulate the real-world situation, and mainly report the 1:N identification performance for measuring the privacy protection rate. Specifically, the 1:N identification aims to identify the image with the same identity as the probe image in the gallery set. Given a probe image and a gallery containing one photo of the same person, the algorithm rank-orders
all the photos in the gallery based on the feature distances to the probe. We test each of the M images per person by adding it to the gallery of distractors and use each of the other M-1 images as a probe. Note that we apply the privacy masks to the probe images, not the gallery images, since we assume that some unauthorized face recognition services may have already obtained the original images of the regular users. In this way, new images cannot be recognized and analyzed, since a wrong image in the gallery set may be matched with the mask-protected probe image. 

We report the Top-1 and Top-5 protection success rate (100\% - Top1 or Top 5 accuracy), which means that the Top-1 and Top-5 images do not have the same identity as the probe. Briefly, higher protection success rate is better. Note that, the more easily a face image is recognized, the more difficult it is to be protected. If a person cannot be recognized, there is no need for us to use a privacy mask to protect it anymore. Therefore, face images in the test dataset should be recognized normally by face recognition models; that is, the initial protection success rate should be almost zero without privacy masks. 

\textbf{Privacy-Commons dataset.} We select 500 persons in the MegaFace challenge2~\cite{nech2017level} database, who have at least 15 clean images (without label noise). For each person, we use at most 10 images as training images to generate the adversarial privacy mask; while another 5 images are used as the testing images to evaluate the protection performance. For the 1:N identification, the selected 500 persons with 5 test images are used as the probe set. Correspondingly, we conduct $500\times5\times4=10,000$ tests. 10,000 distractors, \emph{i.e}., individuals who are not in the probe set, serve as other persons in the gallery set. Note that distractors belong to different persons from the training databases (therefore, different from the 500 persons in the probe set).

\textbf{Privacy-Celebrities dataset.} We select 500 persons in the one-million celebrity list, avoiding repetition of the  MS-Celeb-1M~\cite{Guo16MS} and LFW database~\cite{LFWTech}. For each person, there are at least 20 clean images (without label noise). We use at most 10 images as training images to generate the adversarial privacy mask, while another 10 images are used as the testing images to evaluate the protection performance. For the 1:N identification, the selected 500 persons with 10 test images are used as the probe set. Correspondingly, we conduct $500\times10\times9=45,000$ tests. To reduce the domain gap between probe and gallery set, we use a celebrity database, 13,233 images of LFW database~\cite{LFWTech}, as distractors, which serve as other persons in the gallery set.  

\subsubsection{Face Recognition Models} To simulate potential scenarios, we increase the difference between the source (training) and target (testing) models by attacking deep face recognition models with different training loss functions and network architectures. 

Three source (training) models to generate privacy masks are the modified version~\cite{deng2019arcface} of ResNet-50~\cite{he2016deep} trained on the CAISA-WebFace database~\cite{Yi2014CASIA}, supervised by Softmax loss, CosFace~\cite{Wang2018CosFace}, and ArcFace~\cite{deng2019arcface}. The aim of privacy protection masks is to protect the original face images from being recognized by black-box models. Here, we use six target (testing) models different from the source models. Three of the black-box models differ in loss functions, \emph{i.e}., CosFace~\cite{Wang2018CosFace}, ArcFace~\cite{deng2019arcface} and SFace~\cite{zhong2021sface}. The other three of the black-box models differ in network architectures, \emph{i.e}., the modified version~\cite{deng2019arcface} of the squeeze-and-excitation network (SENet)~\cite{hu2018squeeze}, MobileNet~\cite{howard2017mobilenets}, and Inception-ResNet~\cite{szegedy2017inception}. The recognition performance of the source and black-box models can be found in the Appendix.

\subsection{Protection Performance towards Black-box Face Recognition Models}
\subsubsection{Effectiveness of OPOM}
We evaluate the proposed OPOM method, including OPOM-AffineHull, OPOM-ClassCenter, and OPOM-ConvexHull, on the Privacy-Commons dataset and Privacy-Celebrities dataset. In addition, we also implement the comparison methods, \emph{i.e}., GD-UAP~\cite{mopuri2018generalizable}, FI-UAP, FI-UAP+, FI-UAP-all, and GAP~\cite{poursaeed2018generative} as described in Section~\ref{sec:compare}, to explore their potential capabilities in this person-specific privacy protection task. 

Some details are as follows. The maximum deviation of perturbations $\varepsilon$ is set to 8 under the $L_\infty$ constraint. For GD-UAP, we optimize all the residual blocks at the last layers and the independent convolutional layers following the original paper~\cite{mopuri2018generalizable}. Correspondingly, the number of iterations for the FI-UAP, FI-UAP+, FI-UAP-all and OPOM methods are chosen to be 16, the maximum number of iterations for GD-UAP method is set as 10,000. For GAP, all the images of training datasets are applied to train the Generator for 10 epochs. 

The Top-1 and Top-5 protection success rates (\%) for the 1:N identification on the Privacy-Commons dataset are reported in Table~\ref{table:table_common}. The protection success rates on the Privacy-Celebrities dataset are shown in Table~\ref{table:table_Celebrities}. We can see that OPOM is a more appropriate method for the person-specific privacy protection task. Some protected images with masks generated by OPOM are shown in Figure~\ref{fig:pro}. 

Note that FI-UAP can also be understood as a special case for Equation~\ref{equ:convhull_final}, where $\alpha_i^k=1$ and $\alpha_j^k=0, j \ne i$ for $A_i^k$. FI-UAP+ can be considered as a similar method as OPOM-ClassCenter. From the comparison of FI-UAP, FI-UAP+, OPOM-AffineHull, OPOM-ClassCenter and OPOM-ConvexHull, we can conclude that, the approximation method for describing the feature subspace has influence on the person-specific privacy protection task. FI-UAP only uses the single training point, which cannot make use of other images of this identity; OPOM-AffineHull increases the source space, while it may lead to too loose description; OPOM-ClassCenter (similar to FI-UAP+) indeed relies evenly on all the points of this identity, while it may neglect the differences of the training points $f( {X_{i}^{k}+\Delta {X}} )$; OPOM-ConvexHull increases the source space to an appropriate degree, which will rely on support points differently for a specific training point $f( {X_{i}^{k}+\Delta {X}} )$, and therefore can effectively adapt to different face images better.

\begin{table*}[]
	\center
	\renewcommand\arraystretch{1.0}
	\caption{Comparison of different methods combined with the momentum boosting method~\cite{dong2018boosting} and DFANet~\cite{zhong2020towards} to generate more transferable person-specific privacy masks ($\varepsilon = 8$) from a single source model to protect face images against black-box models. We report Top-1 and Top-5 protection success rate (\%) under 1:N identification setting of the Privacy-Commons dataset. The increment compared with TABLE 2 is indicated by symbol $\uparrow$. }
	\label{table:table_dfanet_common}
	\scalebox{0.98}{
		\begin{tabular}{@{}c|c|cccccccccccc@{}}
			\toprule
			\multirow{3}{*}{\textbf{Source}} & \multirow{3}{*}{\textbf{Method}} & \multicolumn{12}{c}{\textbf{Target}}\\
			&                                  & \multicolumn{2}{c|}{\textbf{ArcFace}}                    & \multicolumn{2}{c|}{\textbf{CosFace}}                    & \multicolumn{2}{c|}{\textbf{SFace}}                      & \multicolumn{2}{c|}{\textbf{MobileNet}}                  & \multicolumn{2}{c|}{\textbf{SENet}}                      & \multicolumn{2}{c}{\textbf{Inception-ResNet}} \\
			&                                  & \textbf{Top-1}   & \multicolumn{1}{c|}{\textbf{Top-5}}   & \textbf{Top-1}   & \multicolumn{1}{c|}{\textbf{Top-5}}   & \textbf{Top-1}   & \multicolumn{1}{c|}{\textbf{Top-5}}   & \textbf{Top-1}   & \multicolumn{1}{c|}{\textbf{Top-5}}   & \textbf{Top-1}   & \multicolumn{1}{c|}{\textbf{Top-5}}   & \textbf{Top-1}        & \textbf{Top-5}        \\ \midrule
			
			\multirow{8}{*}{Softmax}  
			       
			& \multirow{2}{*}{FI-UAP}			 
			& 80.2 & \multicolumn{1}{c|}{71.2}          
			& 72.0 & \multicolumn{1}{c|}{63.9}          
			& 77.9 & \multicolumn{1}{c|}{70.9}          
			& 82.7 & \multicolumn{1}{c|}{72.9}
			& 84.5 & \multicolumn{1}{c|}{77.2}          
			& 61.5 & 50.9 \\
			
			& 
			& \scriptsize\textit{(7.9$\uparrow$)} & \multicolumn{1}{c|}{\scriptsize\textit{(8.9$\uparrow$)}} 
			& \scriptsize\textit{(8.6$\uparrow$)} & \multicolumn{1}{c|}{\scriptsize\textit{(10.5$\uparrow$)}}          
			& \scriptsize\textit{(7.6$\uparrow$)}& \multicolumn{1}{c|}{\scriptsize\textit{(9.1$\uparrow$)}}          
			& \scriptsize\textit{(8.8$\uparrow$)} & \multicolumn{1}{c|}{\scriptsize\textit{(11.2$\uparrow$)}}          
			& \scriptsize\textit{(7.1$\uparrow$)} & \multicolumn{1}{c|}{\scriptsize\textit{(9.6$\uparrow$)}} 
			& \scriptsize\textit{(9.1$\uparrow$)}& \scriptsize\textit{(10.3$\uparrow$)} \\  
			
			& \multirow{2}{*}{OPOM-AffineHull}			 
			& 81.0 & \multicolumn{1}{c|}{72.7}          
			& 73.8 & \multicolumn{1}{c|}{65.3}          
			& 79.1 & \multicolumn{1}{c|}{72.5}          
			& 83.0 & \multicolumn{1}{c|}{73.7}          
			& 85.0 & \multicolumn{1}{c|}{77.9}          
			& 61.7 & 51.6 \\
			
			& 
			& \scriptsize\textit{(7.9$\uparrow$)} & \multicolumn{1}{c|}{\scriptsize\textit{(9.6$\uparrow$)}} 
			& \scriptsize\textit{(10.6$\uparrow$)} & \multicolumn{1}{c|}{\scriptsize\textit{(11.4$\uparrow$)}}          
			& \scriptsize\textit{(8.0$\uparrow$)} & \multicolumn{1}{c|}{\scriptsize\textit{(10.5$\uparrow$)}}          
			& \scriptsize\textit{(8.2$\uparrow$)} & \multicolumn{1}{c|}{\scriptsize\textit{(10.6$\uparrow$)}}          
			& \scriptsize\textit{(7.2$\uparrow$)} & \multicolumn{1}{c|}{\scriptsize\textit{(9.3$\uparrow$)}} 
			& \scriptsize\textit{(8.9$\uparrow$)} & \scriptsize\textit{(10.8$\uparrow$)} \\  
			
			& \multirow{2}{*}{OPOM-ClassCenter}			 
			& 80.6 & \multicolumn{1}{c|}{72.2} 
			& 73.9 & \multicolumn{1}{c|}{65.6}          
			& 78.8 & \multicolumn{1}{c|}{72.1} 
			& 83.1 & \multicolumn{1}{c|}{74.0}          
			& 85.0 & \multicolumn{1}{c|}{78.2}          
			& 62.5 & 51.9 \\
			
			& 
			& \scriptsize\textit{(3.7$\uparrow$)} & \multicolumn{1}{c|}{\scriptsize\textit{(4.4$\uparrow$)}} 
			& \scriptsize\textit{(4.7$\uparrow$)} & \multicolumn{1}{c|}{\scriptsize\textit{(5.3$\uparrow$)}}          
			& \scriptsize\textit{(3.9$\uparrow$)}& \multicolumn{1}{c|}{\scriptsize\textit{(4.8$\uparrow$)}}          
			& \scriptsize\textit{(4.6$\uparrow$)} & \multicolumn{1}{c|}{\scriptsize\textit{(6.1$\uparrow$)}}          
			& \scriptsize\textit{(3.0$\uparrow$)} & \multicolumn{1}{c|}{\scriptsize\textit{(4.9$\uparrow$)}} 
			& \scriptsize\textit{(5.4$\uparrow$)}& \scriptsize\textit{(6.4$\uparrow$)} \\  
			
			& \multirow{2}{*}{OPOM-ConvexHull}			 
			& 81.2 & \multicolumn{1}{c|}{72.6}          
			& 73.6 & \multicolumn{1}{c|}{65.7}          
			& 79.1 & \multicolumn{1}{c|}{72.2}          
			& 83.6 & \multicolumn{1}{c|}{74.6}          
			& 85.0 & \multicolumn{1}{c|}{78.0}          
			& 62.4 & 52.1 \\

			& 
			& \scriptsize\textit{(3.3$\uparrow$)} & \multicolumn{1}{c|}{\scriptsize\textit{(3.3$\uparrow$)}} 
			& \scriptsize\textit{(3.4$\uparrow$)} & \multicolumn{1}{c|}{\scriptsize\textit{(4.3$\uparrow$)}}          
			& \scriptsize\textit{(3.0$\uparrow$)}& \multicolumn{1}{c|}{\scriptsize\textit{(3.5$\uparrow$)}}          
			& \scriptsize\textit{(4.4$\uparrow$)} & \multicolumn{1}{c|}{\scriptsize\textit{(5.5$\uparrow$)}}          
			& \scriptsize\textit{(2.1$\uparrow$)} & \multicolumn{1}{c|}{\scriptsize\textit{(3.8$\uparrow$)}} 
			& \scriptsize\textit{(3.7$\uparrow$)}& \scriptsize\textit{(4.8$\uparrow$)} \\  
			
			\midrule
			
			\multirow{8}{*}{ArcFace}         			
			
			& \multirow{2}{*}{FI-UAP}			 
			& 89.8 & \multicolumn{1}{c|}{84.9} 
			& 81.1 & \multicolumn{1}{c|}{75.3}
			& 86.2 & \multicolumn{1}{c|}{81.0}
			& 79.3 & \multicolumn{1}{c|}{68.5}          
			& 85.6 & \multicolumn{1}{c|}{78.8}          
			& 70.6 & 60.9 \\
			
			& 
			& \scriptsize\textit{(7.5$\uparrow$)} & \multicolumn{1}{c|}{\scriptsize\textit{(9.9$\uparrow$)}} 
			& \scriptsize\textit{(9.9$\uparrow$)} & \multicolumn{1}{c|}{\scriptsize\textit{(11.7$\uparrow$)}}          
			& \scriptsize\textit{(8.9$\uparrow$)}& \multicolumn{1}{c|}{\scriptsize\textit{(11.0$\uparrow$)}}          
			& \scriptsize\textit{(14.1$\uparrow$)} & \multicolumn{1}{c|}{\scriptsize\textit{(18.2$\uparrow$)}}          
			& \scriptsize\textit{(11.7$\uparrow$)} & \multicolumn{1}{c|}{\scriptsize\textit{(15.7$\uparrow$)}} 
			& \scriptsize\textit{(14.0$\uparrow$)}& \scriptsize\textit{(15.5$\uparrow$)} \\  
			
			& \multirow{2}{*}{OPOM-AffineHull}			 
			& 91.0 & \multicolumn{1}{c|}{86.1} 
			& 82.9 & \multicolumn{1}{c|}{77.3}          
			& 87.5 & \multicolumn{1}{c|}{82.7}          
			& 80.9 & \multicolumn{1}{c|}{70.9}          
			& 86.4 & \multicolumn{1}{c|}{80.5}          
			& 72.1 & 63.4 \\
			
			& 
			& \scriptsize\textit{(8.1$\uparrow$)} & \multicolumn{1}{c|}{\scriptsize\textit{(10.3$\uparrow$)}} 
			& \scriptsize\textit{(10.6$\uparrow$)} & \multicolumn{1}{c|}{\scriptsize\textit{(12.6$\uparrow$)}}          
			& \scriptsize\textit{(9.4$\uparrow$)}& \multicolumn{1}{c|}{\scriptsize\textit{(11.8$\uparrow$)}}          
			& \scriptsize\textit{(14.6$\uparrow$)} & \multicolumn{1}{c|}{\scriptsize\textit{(18.7$\uparrow$)}}          
			& \scriptsize\textit{(11.0$\uparrow$)} & \multicolumn{1}{c|}{\scriptsize\textit{(16.1$\uparrow$)}} 
			& \scriptsize\textit{(13.7$\uparrow$)}& \scriptsize\textit{(16.6$\uparrow$)} \\  
			
			& \multirow{2}{*}{OPOM-ClassCenter}			 
			& 90.7 & \multicolumn{1}{c|}{86.1}          
			& 82.9 & \multicolumn{1}{c|}{77.1}          
			& 87.6 & \multicolumn{1}{c|}{82.5}          
			& 80.6 & \multicolumn{1}{c|}{70.5} 
			& 86.5 & \multicolumn{1}{c|}{80.1}          
			& 72.1 & 62.4 \\
			
			& 
			& \scriptsize\textit{(4.8$\uparrow$)} & \multicolumn{1}{c|}{\scriptsize\textit{(6.6$\uparrow$)}} 
			& \scriptsize\textit{(6.8$\uparrow$)} & \multicolumn{1}{c|}{\scriptsize\textit{(7.5$\uparrow$)}}          
			& \scriptsize\textit{(5.6$\uparrow$)}& \multicolumn{1}{c|}{\scriptsize\textit{(6.7$\uparrow$)}}          
			& \scriptsize\textit{(10.9$\uparrow$)} & \multicolumn{1}{c|}{\scriptsize\textit{(13.8$\uparrow$)}}          
			& \scriptsize\textit{(8.0$\uparrow$)} & \multicolumn{1}{c|}{\scriptsize\textit{(10.4$\uparrow$)}} 
			& \scriptsize\textit{(10.3$\uparrow$)}& \scriptsize\textit{(11.4$\uparrow$)} \\  
			
			& \multirow{2}{*}{OPOM-ConvexHull}			 
			& 90.7 & \multicolumn{1}{c|}{86.2}          
			& 82.7 & \multicolumn{1}{c|}{77.0}          
			& 87.6 & \multicolumn{1}{c|}{82.7}          
			& 81.2 & \multicolumn{1}{c|}{71.3}          
			& 86.5 & \multicolumn{1}{c|}{80.6}          
			& 72.0 & 63.0 \\
			
			& 
			& \scriptsize\textit{(4.3$\uparrow$)} & \multicolumn{1}{c|}{\scriptsize\textit{(6.0$\uparrow$)}} 
			& \scriptsize\textit{(5.9$\uparrow$)} & \multicolumn{1}{c|}{\scriptsize\textit{(7.1$\uparrow$)}}          
			& \scriptsize\textit{(4.9$\uparrow$)}& \multicolumn{1}{c|}{\scriptsize\textit{(6.2$\uparrow$)}}          
			& \scriptsize\textit{(10.7$\uparrow$)} & \multicolumn{1}{c|}{\scriptsize\textit{(14.0$\uparrow$)}}          
			& \scriptsize\textit{(7.2$\uparrow$)} & \multicolumn{1}{c|}{\scriptsize\textit{(10.4$\uparrow$)}} 
			& \scriptsize\textit{(8.8$\uparrow$)}& \scriptsize\textit{(10.8$\uparrow$)} \\  
			\midrule
			
			\multirow{8}{*}{CosFace}        
			
			& \multirow{2}{*}{FI-UAP}			 
			& 88.6 & \multicolumn{1}{c|}{82.8}          
			& 82.5 & \multicolumn{1}{c|}{76.4}          
			& 85.5 & \multicolumn{1}{c|}{79.7}          
			& 69.7 & \multicolumn{1}{c|}{56.0}          
			& 79.2 & \multicolumn{1}{c|}{69.6}          
			& 62.0 & 51.3  \\
			
			& 
			& \scriptsize\textit{(8.2$\uparrow$)} & \multicolumn{1}{c|}{\scriptsize\textit{(10.5$\uparrow$)}} 
			& \scriptsize\textit{(9.7$\uparrow$)} & \multicolumn{1}{c|}{\scriptsize\textit{(11.2$\uparrow$)}}          
			& \scriptsize\textit{(8.9$\uparrow$)}& \multicolumn{1}{c|}{\scriptsize\textit{(10.8$\uparrow$)}}          
			& \scriptsize\textit{(14.1$\uparrow$)} & \multicolumn{1}{c|}{\scriptsize\textit{(15.0$\uparrow$)}}          
			& \scriptsize\textit{(12.3$\uparrow$)} & \multicolumn{1}{c|}{\scriptsize\textit{(15.1$\uparrow$)}} 
			& \scriptsize\textit{(12.0$\uparrow$)}& \scriptsize\textit{(14.2$\uparrow$)} \\  
			
			& \multirow{2}{*}{OPOM-AffineHull}			 
			& 89.4 & \multicolumn{1}{c|}{84.3}          
			& 84.0 & \multicolumn{1}{c|}{78.8}          
			& 86.9 & \multicolumn{1}{c|}{81.8}          
			& 72.9 & \multicolumn{1}{c|}{60.7}          
			& 81.8 & \multicolumn{1}{c|}{73.4}          
			& 64.5 & 53.8 \\
			
			& 
			& \scriptsize\textit{(7.1$\uparrow$)} & \multicolumn{1}{c|}{\scriptsize\textit{(9.8$\uparrow$)}} 
			& \scriptsize\textit{(9.7$\uparrow$)} & \multicolumn{1}{c|}{\scriptsize\textit{(11.8$\uparrow$)}}          
			& \scriptsize\textit{(8.4$\uparrow$)}& \multicolumn{1}{c|}{\scriptsize\textit{(10.5$\uparrow$)}}          
			& \scriptsize\textit{(14.3$\uparrow$)} & \multicolumn{1}{c|}{\scriptsize\textit{(16.9$\uparrow$)}}          
			& \scriptsize\textit{(12.8$\uparrow$)} & \multicolumn{1}{c|}{\scriptsize\textit{(16.6$\uparrow$)}} 
			& \scriptsize\textit{(12.7$\uparrow$)}& \scriptsize\textit{(14.4$\uparrow$)} \\  
			
			& \multirow{2}{*}{OPOM-ClassCenter}			 
			& 89.4 & \multicolumn{1}{c|}{84.1}          
			& 83.8 & \multicolumn{1}{c|}{78.4}          
			& 86.8 & \multicolumn{1}{c|}{81.5}          
			& 72.4 & \multicolumn{1}{c|}{60.4}          
			& 81.1 & \multicolumn{1}{c|}{72.1}          
			& 64.4 & 53.7 \\
			
			& 
			& \scriptsize\textit{(4.0$\uparrow$)} & \multicolumn{1}{c|}{\scriptsize\textit{(5.7$\uparrow$)}} 
			& \scriptsize\textit{(5.0$\uparrow$)} & \multicolumn{1}{c|}{\scriptsize\textit{(6.8$\uparrow$)}}          
			& \scriptsize\textit{(4.4$\uparrow$)}& \multicolumn{1}{c|}{\scriptsize\textit{(6.0$\uparrow$)}}          
			& \scriptsize\textit{(11.0$\uparrow$)} & \multicolumn{1}{c|}{\scriptsize\textit{(12.9$\uparrow$)}}          
			& \scriptsize\textit{(8.1$\uparrow$)} & \multicolumn{1}{c|}{\scriptsize\textit{(10.6$\uparrow$)}} 
			& \scriptsize\textit{(9.2$\uparrow$)}& \scriptsize\textit{(10.8$\uparrow$)} \\  
			
			& \multirow{2}{*}{OPOM-ConvexHull}			 
			& 89.6 & \multicolumn{1}{c|}{84.1}          
			& 84.2 & \multicolumn{1}{c|}{78.8}          
			& 87.8 & \multicolumn{1}{c|}{82.3}          
			& 73.0 & \multicolumn{1}{c|}{60.8}          
			& 82.0 & \multicolumn{1}{c|}{73.4}          
			& 64.8 & 53.6 \\
			
			& 
			& \scriptsize\textit{(3.0$\uparrow$)} & \multicolumn{1}{c|}{\scriptsize\textit{(4.7$\uparrow$)}} 
			& \scriptsize\textit{(4.7$\uparrow$)} & \multicolumn{1}{c|}{\scriptsize\textit{(6.1$\uparrow$)}}          
			& \scriptsize\textit{(4.8$\uparrow$)}& \multicolumn{1}{c|}{\scriptsize\textit{(6.0$\uparrow$)}}          
			& \scriptsize\textit{(10.7$\uparrow$)} & \multicolumn{1}{c|}{\scriptsize\textit{(12.6$\uparrow$)}}          
			& \scriptsize\textit{(8.1$\uparrow$)} & \multicolumn{1}{c|}{\scriptsize\textit{(10.2$\uparrow$)}} 
			& \scriptsize\textit{(8.0$\uparrow$)}& \scriptsize\textit{(9.1$\uparrow$)} \\  
			\bottomrule
	\end{tabular}}
\end{table*}

\begin{table*}[]
	\center
	\renewcommand\arraystretch{1.0}
	\caption{Comparison of different methods combined with the momentum boosting method~\cite{dong2018boosting} and DFANet~\cite{zhong2020towards} to generate more transferable person-specific privacy masks ($\varepsilon = 8$) from a single source model to protect face images against black-box models. We report Top-1 and Top-5 protection success rate (\%) under 1:N identification setting of the Privacy-Celebrities dataset. The increment compared with TABLE 3 is indicated by symbol $\uparrow$.}
	\label{table:table_dfanet_Celebrities}
	\scalebox{0.98}{
		\begin{tabular}{@{}c|c|cccccccccccc@{}}
	\toprule
	\multirow{3}{*}{\textbf{Source}} & \multirow{3}{*}{\textbf{Method}} & \multicolumn{12}{c}{\textbf{Target}}\\
	&                                  & \multicolumn{2}{c|}{\textbf{ArcFace}}                    & \multicolumn{2}{c|}{\textbf{CosFace}}                    & \multicolumn{2}{c|}{\textbf{SFace}}                      & \multicolumn{2}{c|}{\textbf{MobileNet}}                  & \multicolumn{2}{c|}{\textbf{SENet}}                      & \multicolumn{2}{c}{\textbf{Inception-ResNet}} \\
	&                                  & \textbf{Top-1}   & \multicolumn{1}{c|}{\textbf{Top-5}}   & \textbf{Top-1}   & \multicolumn{1}{c|}{\textbf{Top-5}}   & \textbf{Top-1}   & \multicolumn{1}{c|}{\textbf{Top-5}}   & \textbf{Top-1}   & \multicolumn{1}{c|}{\textbf{Top-5}}   & \textbf{Top-1}   & \multicolumn{1}{c|}{\textbf{Top-5}}   & \textbf{Top-1}        & \textbf{Top-5}        \\ \midrule
	
	\multirow{8}{*}{Softmax}  
	
	& \multirow{2}{*}{FI-UAP}			 
	& 62.7 & \multicolumn{1}{c|}{52.2}          
	& 52.9 & \multicolumn{1}{c|}{43.2}          
	& 60.3 & \multicolumn{1}{c|}{50.8}          
	& 67.8 & \multicolumn{1}{c|}{56.2}          
	& 68.6 & \multicolumn{1}{c|}{58.6}          
	& 43.6 & 32.1 \\
	
	& 
	& \scriptsize\textit{(6.2$\uparrow$)} & \multicolumn{1}{c|}{\scriptsize\textit{(6.6$\uparrow$)}} 
	& \scriptsize\textit{(7.7$\uparrow$)} & \multicolumn{1}{c|}{\scriptsize\textit{(8.3$\uparrow$)}}          
	& \scriptsize\textit{(7.4$\uparrow$)}& \multicolumn{1}{c|}{\scriptsize\textit{(8.3$\uparrow$)}}          
	& \scriptsize\textit{(7.6$\uparrow$)} & \multicolumn{1}{c|}{\scriptsize\textit{(9.3$\uparrow$)}}          
	& \scriptsize\textit{(7.0$\uparrow$)} & \multicolumn{1}{c|}{\scriptsize\textit{(8.8$\uparrow$)}} 
	& \scriptsize\textit{(6.2$\uparrow$)}& \scriptsize\textit{(5.6$\uparrow$)} \\  
	
	& \multirow{2}{*}{OPOM-AffineHull}			 
	& 65.1 & \multicolumn{1}{c|}{54.8}          
	& 54.9 & \multicolumn{1}{c|}{45.1}          
	& 62.4 & \multicolumn{1}{c|}{53.2}          
	& 69.9 & \multicolumn{1}{c|}{59.0}          
	& 70.5 & \multicolumn{1}{c|}{60.8}          
	& 45.0 & 33.4 \\
	
	& 
	& \scriptsize\textit{(6.3$\uparrow$)} & \multicolumn{1}{c|}{\scriptsize\textit{(7.0$\uparrow$)}} 
	& \scriptsize\textit{(8.0$\uparrow$)} & \multicolumn{1}{c|}{\scriptsize\textit{(7.9$\uparrow$)}}          
	& \scriptsize\textit{(7.6$\uparrow$)}& \multicolumn{1}{c|}{\scriptsize\textit{(8.3$\uparrow$)}}          
	& \scriptsize\textit{(7.4$\uparrow$)} & \multicolumn{1}{c|}{\scriptsize\textit{(9.6$\uparrow$)}}          
	& \scriptsize\textit{(7.0$\uparrow$)} & \multicolumn{1}{c|}{\scriptsize\textit{(8.6$\uparrow$)}} 
	& \scriptsize\textit{(6.0$\uparrow$)}& \scriptsize\textit{(5.5$\uparrow$)} \\  
	
	& \multirow{2}{*}{OPOM-ClassCenter}			 
	& 64.3 & \multicolumn{1}{c|}{54.0}          
	& 54.2 & \multicolumn{1}{c|}{45.4}          
	& 61.7 & \multicolumn{1}{c|}{52.5}          
	& 69.7 & \multicolumn{1}{c|}{58.7}          
	& 69.8 & \multicolumn{1}{c|}{60.1}          
	& 45.2 & 33.9 \\
	
	& 
	& \scriptsize\textit{(1.9$\uparrow$)} & \multicolumn{1}{c|}{\scriptsize\textit{(2.5$\uparrow$)}} 
	& \scriptsize\textit{(2.36$\uparrow$)} & \multicolumn{1}{c|}{\scriptsize\textit{(3.0$\uparrow$)}}          
	& \scriptsize\textit{(1.9$\uparrow$)}& \multicolumn{1}{c|}{\scriptsize\textit{(2.8$\uparrow$)}}          
	& \scriptsize\textit{(3.5$\uparrow$)} & \multicolumn{1}{c|}{\scriptsize\textit{(3.9$\uparrow$)}}          
	& \scriptsize\textit{(2.1$\uparrow$)} & \multicolumn{1}{c|}{\scriptsize\textit{(2.4$\uparrow$)}} 
	& \scriptsize\textit{(2.8$\uparrow$)}& \scriptsize\textit{(2.5$\uparrow$)} \\  
	
	& \multirow{2}{*}{OPOM-ConvexHull}			 
	& 65.7 & \multicolumn{1}{c|}{55.5}          
	& 55.5 & \multicolumn{1}{c|}{46.0}          
	& 63.0 & \multicolumn{1}{c|}{53.7} 
	& 70.1 & \multicolumn{1}{c|}{59.1} 
	& 70.9 & \multicolumn{1}{c|}{61.3}          
	& 46.5 & 35.3 \\
	
	& 
	& \scriptsize\textit{(1.9$\uparrow$)} & \multicolumn{1}{c|}{\scriptsize\textit{(2.0$\uparrow$)}} 
	& \scriptsize\textit{(1.9$\uparrow$)} & \multicolumn{1}{c|}{\scriptsize\textit{(2.1$\uparrow$)}}          
	& \scriptsize\textit{(1.7$\uparrow$)}& \multicolumn{1}{c|}{\scriptsize\textit{(2.1$\uparrow$)}}          
	& \scriptsize\textit{(3.1$\uparrow$)} & \multicolumn{1}{c|}{\scriptsize\textit{(3.4$\uparrow$)}}          
	& \scriptsize\textit{(2.0$\uparrow$)} & \multicolumn{1}{c|}{\scriptsize\textit{(2.4$\uparrow$)}} 
	& \scriptsize\textit{(2.3$\uparrow$)}& \scriptsize\textit{(2.1$\uparrow$)} \\  
	\midrule
	
	\multirow{8}{*}{ArcFace}         			
	
	& \multirow{2}{*}{FI-UAP}			 
	& 73.0 & \multicolumn{1}{c|}{64.5}          
	& 60.3 & \multicolumn{1}{c|}{50.9} 
	& 67.4 & \multicolumn{1}{c|}{58.9}          
	& 63.9 & \multicolumn{1}{c|}{50.4}          
	& 68.0 & \multicolumn{1}{c|}{56.8}          
	& 49.2 & 37.6\\
	
	& 
	& \scriptsize\textit{(10.5$\uparrow$)} & \multicolumn{1}{c|}{\scriptsize\textit{(13.1$\uparrow$)}} 
	& \scriptsize\textit{(12.4$\uparrow$)} & \multicolumn{1}{c|}{\scriptsize\textit{(13.1$\uparrow$)}}          
	& \scriptsize\textit{(11.8$\uparrow$)}& \multicolumn{1}{c|}{\scriptsize\textit{(13.0$\uparrow$)}}          
	& \scriptsize\textit{(13.9$\uparrow$)} & \multicolumn{1}{c|}{\scriptsize\textit{(14.8$\uparrow$)}}          
	& \scriptsize\textit{(13.7$\uparrow$)} & \multicolumn{1}{c|}{\scriptsize\textit{(15.8$\uparrow$)}} 
	& \scriptsize\textit{(11.5$\uparrow$)}& \scriptsize\textit{(11.3$\uparrow$)} \\  
	
	& \multirow{2}{*}{OPOM-AffineHull}			 
	& 75.9 & \multicolumn{1}{c|}{67.7}          
	& 64.0 & \multicolumn{1}{c|}{55.4}          
	& 71.0 & \multicolumn{1}{c|}{63.4}          
	& 67.0 & \multicolumn{1}{c|}{54.8}          
	& 71.5 & \multicolumn{1}{c|}{61.1}          
	& 53.1 & 41.6 \\
	
	& 
	& \scriptsize\textit{(10.1$\uparrow$)} & \multicolumn{1}{c|}{\scriptsize\textit{(12.5$\uparrow$)}} 
	& \scriptsize\textit{(12.7$\uparrow$)} & \multicolumn{1}{c|}{\scriptsize\textit{(14.1$\uparrow$)}}          
	& \scriptsize\textit{(11.9$\uparrow$)}& \multicolumn{1}{c|}{\scriptsize\textit{(13.8$\uparrow$)}}          
	& \scriptsize\textit{(13.9$\uparrow$)} & \multicolumn{1}{c|}{\scriptsize\textit{(15.7$\uparrow$)}}          
	& \scriptsize\textit{(13.8$\uparrow$)} & \multicolumn{1}{c|}{\scriptsize\textit{(16.0$\uparrow$)}} 
	& \scriptsize\textit{(12.6$\uparrow$)}& \scriptsize\textit{(13.2$\uparrow$)} \\  
	
	& \multirow{2}{*}{OPOM-ClassCenter}			 
	& 75.3 & \multicolumn{1}{c|}{67.2}          
	& 63.4 & \multicolumn{1}{c|}{54.9}          
	& 70.6 & \multicolumn{1}{c|}{62.8}          
	& 66.0 & \multicolumn{1}{c|}{53.5}          
	& 70.8 & \multicolumn{1}{c|}{60.4}          
	& 52.4 & 40.9 \\
	
	& 
	& \scriptsize\textit{(6.5$\uparrow$)} & \multicolumn{1}{c|}{\scriptsize\textit{(7.9$\uparrow$)}} 
	& \scriptsize\textit{(8.2$\uparrow$)} & \multicolumn{1}{c|}{\scriptsize\textit{(8.8$\uparrow$)}}          
	& \scriptsize\textit{(7.6$\uparrow$)}& \multicolumn{1}{c|}{\scriptsize\textit{(8.5$\uparrow$)}}          
	& \scriptsize\textit{(9.4$\uparrow$)} & \multicolumn{1}{c|}{\scriptsize\textit{(10.9$\uparrow$)}}          
	& \scriptsize\textit{(9.7$\uparrow$)} & \multicolumn{1}{c|}{\scriptsize\textit{(11.4$\uparrow$)}} 
	& \scriptsize\textit{(8.5$\uparrow$)}& \scriptsize\textit{(8.9$\uparrow$)} \\  
	
	& \multirow{2}{*}{OPOM-ConvexHull}			 
	& 76.3 & \multicolumn{1}{c|}{68.0}          
	& 64.4 & \multicolumn{1}{c|}{55.6}          
	& 71.2 & \multicolumn{1}{c|}{63.4}          
	& 67.0 & \multicolumn{1}{c|}{54.5}          
	& 72.0 & \multicolumn{1}{c|}{61.4}          
	& 53.9 & 42.3\\
	
	& 
	& \scriptsize\textit{(6.5$\uparrow$)} & \multicolumn{1}{c|}{\scriptsize\textit{(7.6$\uparrow$)}} 
	& \scriptsize\textit{(8.2$\uparrow$)} & \multicolumn{1}{c|}{\scriptsize\textit{(8.3$\uparrow$)}}          
	& \scriptsize\textit{(7.1$\uparrow$)}& \multicolumn{1}{c|}{\scriptsize\textit{(8.2$\uparrow$)}}          
	& \scriptsize\textit{(9.5$\uparrow$)} & \multicolumn{1}{c|}{\scriptsize\textit{(11.4$\uparrow$)}}          
	& \scriptsize\textit{(9.7$\uparrow$)} & \multicolumn{1}{c|}{\scriptsize\textit{(11.4$\uparrow$)}} 
	& \scriptsize\textit{(9.0$\uparrow$)}& \scriptsize\textit{(9.1$\uparrow$)} \\  
	\midrule
	
	\multirow{8}{*}{CosFace}        
	& \multirow{2}{*}{FI-UAP}			 
	& 71.7 & \multicolumn{1}{c|}{62.1}          
	& 60.7 & \multicolumn{1}{c|}{51.1}          
	& 66.5 & \multicolumn{1}{c|}{57.4}          
	& 56.0 & \multicolumn{1}{c|}{41.7}          
	& 61.4 & \multicolumn{1}{c|}{48.9}          
	& 42.4 & 30.3 \\
	
	& 
	& \scriptsize\textit{(8.0$\uparrow$)} & \multicolumn{1}{c|}{\scriptsize\textit{(10.1$\uparrow$)}} 
	& \scriptsize\textit{(9.1$\uparrow$)} & \multicolumn{1}{c|}{\scriptsize\textit{(9.7$\uparrow$)}}          
	& \scriptsize\textit{(8.8$\uparrow$)}& \multicolumn{1}{c|}{\scriptsize\textit{(10.2$\uparrow$)}}          
	& \scriptsize\textit{(9.5$\uparrow$)} & \multicolumn{1}{c|}{\scriptsize\textit{(10.0$\uparrow$)}}          
	& \scriptsize\textit{(10.6$\uparrow$)} & \multicolumn{1}{c|}{\scriptsize\textit{(11.9$\uparrow$)}} 
	& \scriptsize\textit{(9.4$\uparrow$)}& \scriptsize\textit{(8.5$\uparrow$)} \\  
	
	& \multirow{2}{*}{OPOM-AffineHull}			 
	& 74.9 & \multicolumn{1}{c|}{66.1}          
	& 65.1 & \multicolumn{1}{c|}{56.8}          
	& 70.6 & \multicolumn{1}{c|}{62.7}          
	& 60.7 & \multicolumn{1}{c|}{46.5}          
	& 66.2 & \multicolumn{1}{c|}{54.9}          
	& 46.2 & 34.4 \\
	
	& 
	& \scriptsize\textit{(8.4$\uparrow$)} & \multicolumn{1}{c|}{\scriptsize\textit{(10.5$\uparrow$)}} 
	& \scriptsize\textit{(9.6$\uparrow$)} & \multicolumn{1}{c|}{\scriptsize\textit{(11.4$\uparrow$)}}          
	& \scriptsize\textit{(9.5$\uparrow$)}& \multicolumn{1}{c|}{\scriptsize\textit{(11.2$\uparrow$)}}          
	& \scriptsize\textit{(11.2$\uparrow$)} & \multicolumn{1}{c|}{\scriptsize\textit{(11.8$\uparrow$)}}          
	& \scriptsize\textit{(11.5$\uparrow$)} & \multicolumn{1}{c|}{\scriptsize\textit{(14.1$\uparrow$)}} 
	& \scriptsize\textit{(10.5$\uparrow$)}& \scriptsize\textit{(10.5$\uparrow$)} \\  
	
	& \multirow{2}{*}{OPOM-ClassCenter}			 
	& 73.8 & \multicolumn{1}{c|}{64.8}          
	& 64.3 & \multicolumn{1}{c|}{55.9}          
	& 69.8 & \multicolumn{1}{c|}{61.6}          
	& 59.9 & \multicolumn{1}{c|}{45.3}          
	& 65.1 & \multicolumn{1}{c|}{53.4}          
	& 45.7 & 33.6 \\
	
	& 
	& \scriptsize\textit{(5.5$\uparrow$)} & \multicolumn{1}{c|}{\scriptsize\textit{(6.4$\uparrow$)}} 
	& \scriptsize\textit{(5.8$\uparrow$)} & \multicolumn{1}{c|}{\scriptsize\textit{(7.1$\uparrow$)}}          
	& \scriptsize\textit{(6.0$\uparrow$)}& \multicolumn{1}{c|}{\scriptsize\textit{(7.1$\uparrow$)}}          
	& \scriptsize\textit{(8.7$\uparrow$)} & \multicolumn{1}{c|}{\scriptsize\textit{(9.0$\uparrow$)}}          
	& \scriptsize\textit{(8.1$\uparrow$)} & \multicolumn{1}{c|}{\scriptsize\textit{(9.5$\uparrow$)}} 
	& \scriptsize\textit{(7.2$\uparrow$)}& \scriptsize\textit{(7.1$\uparrow$)} \\  
	
	& \multirow{2}{*}{OPOM-ConvexHull}			 
	& 75.2 & \multicolumn{1}{c|}{66.6}          
	& 65.1 & \multicolumn{1}{c|}{57.0}          
	& 70.9 & \multicolumn{1}{c|}{63.0}          
	& 60.6 & \multicolumn{1}{c|}{46.4}          
	& 66.2 & \multicolumn{1}{c|}{54.9}          
	& 47.0 & 34.8 \\
	
	& 
	& \scriptsize\textit{(5.3$\uparrow$)} & \multicolumn{1}{c|}{\scriptsize\textit{(6.0$\uparrow$)}} 
	& \scriptsize\textit{(5.3$\uparrow$)} & \multicolumn{1}{c|}{\scriptsize\textit{(6.5$\uparrow$)}}          
	& \scriptsize\textit{(5.3$\uparrow$)}& \multicolumn{1}{c|}{\scriptsize\textit{(6.8$\uparrow$)}}          
	& \scriptsize\textit{(7.6$\uparrow$)} & \multicolumn{1}{c|}{\scriptsize\textit{(8.3$\uparrow$)}}          
	& \scriptsize\textit{(7.5$\uparrow$)} & \multicolumn{1}{c|}{\scriptsize\textit{(9.4$\uparrow$)}} 
	& \scriptsize\textit{(6.8$\uparrow$)}& \scriptsize\textit{(6.7$\uparrow$)} \\  
	\bottomrule
	\end{tabular}}
\end{table*}

\subsubsection{Combination with Transferability Methods}
As privacy protection masks should protect face images from different kinds of unknown models, it is vital to generate transferable masks. We investigate the performance of the proposed OPOM combined with transferability enhancement methods including the momentum boosting method~\cite{dong2018boosting} and DFANet~\cite{zhong2020towards}, as mentioned in Section~\ref{sec:trans}. Experimental results on the Privacy-Commons dataset and Privacy-Celebrities dataset are listed in Table~\ref{table:table_dfanet_common} and Table~\ref{table:table_dfanet_Celebrities} respectively. With the enhancement of the momentum boosting method and DFANet, the privacy protection success rate of OPOM can increase further, while there still exists some scope for improvement.

\begin{figure}[htbp]
	\center
	\includegraphics[width=1\linewidth]{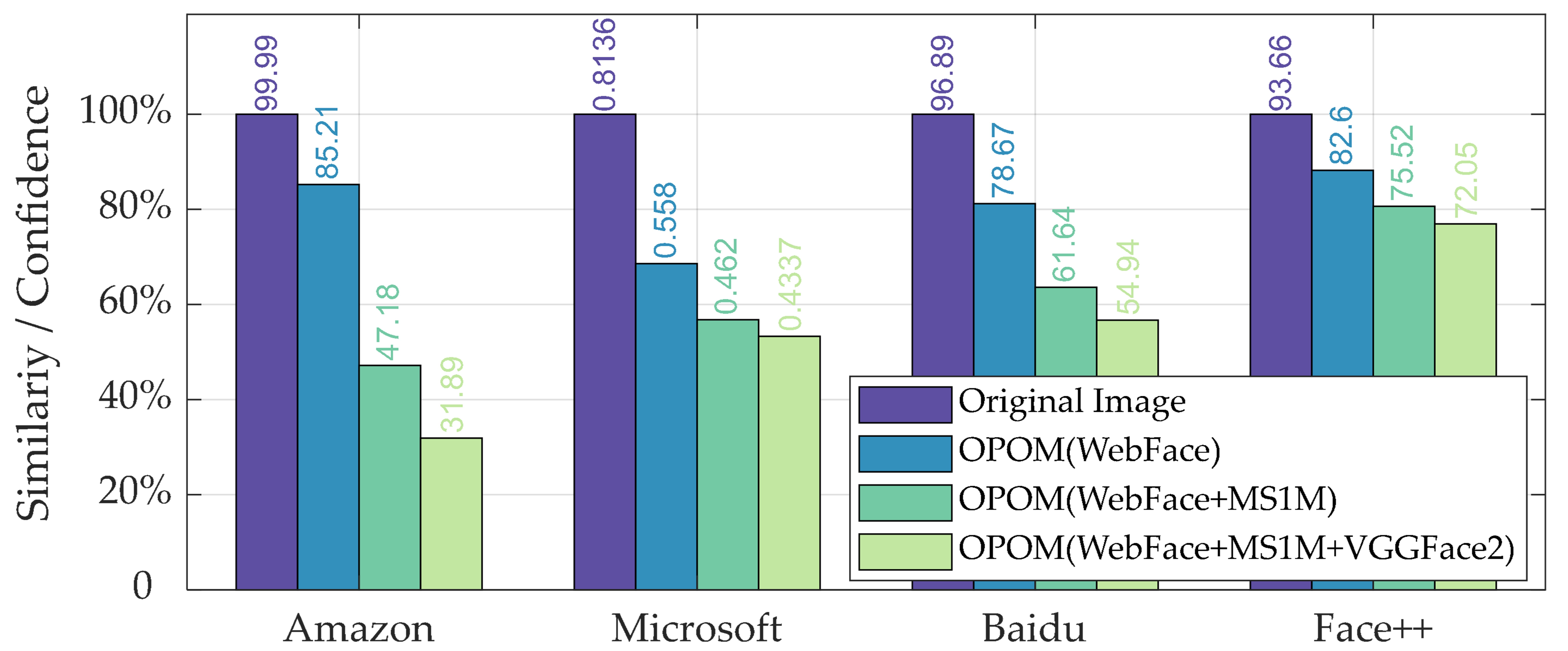}
	\caption{Protection against Commercial APIs (Amazon~\cite{Amazon}, Microsoft~\cite{azure}, Baidu~\cite{Baidu} and Face++~\cite{Face++}). Fifty identities in the Privacy-Commons dataset, each with 5 test images are used for the face verification test. The normalized average similarity/confidence scores are shown (lower is better). The original scores are listed above the bar.}
	\label{fig:API}
\end{figure}

\begin{figure}[htbp]
	\center
	\includegraphics[width=1\linewidth]{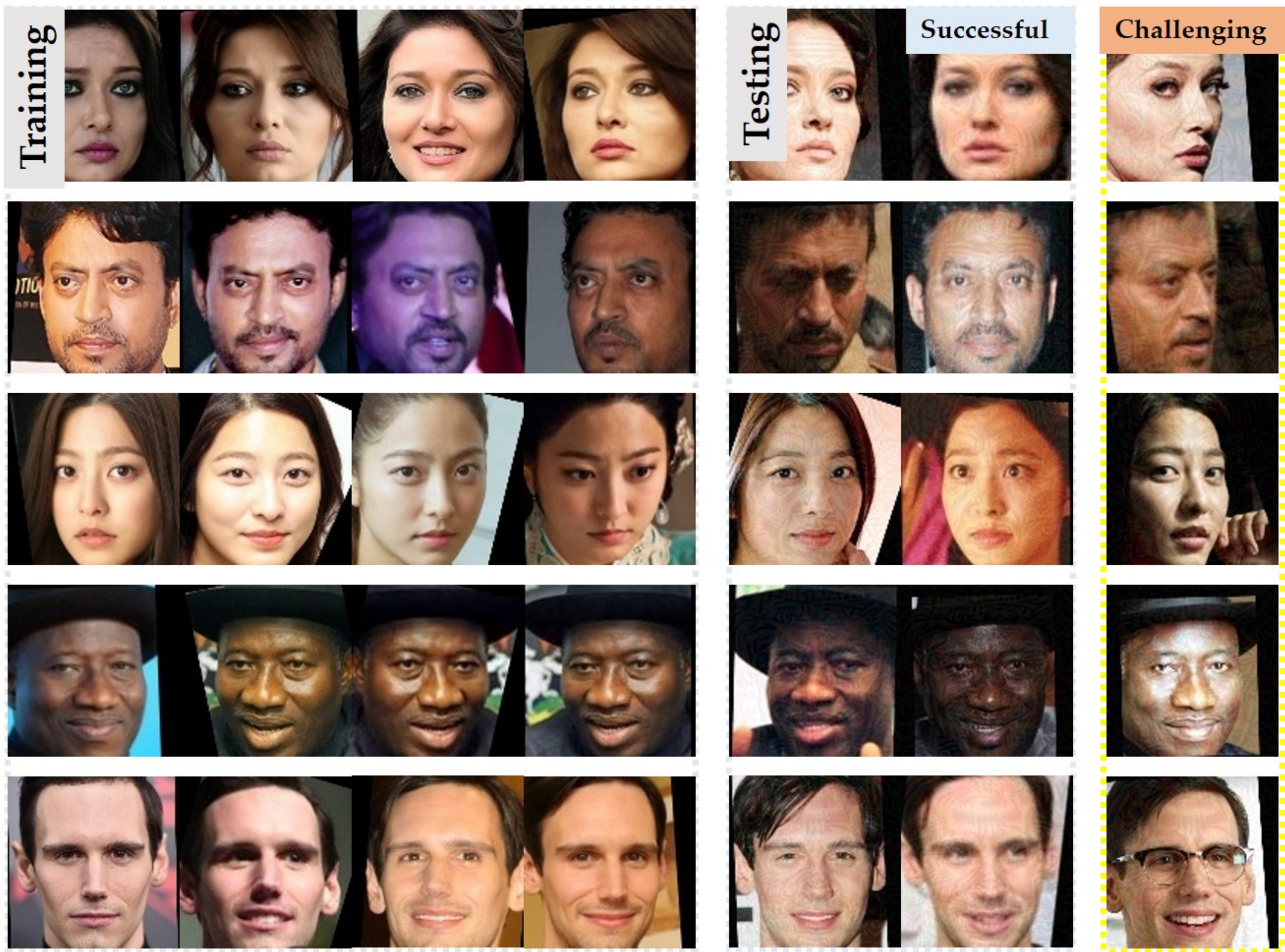}
	\caption{Some failure cases, as well as the corresponding training samples and successful easily protected samples for analysis. Each row represents an identity. The privacy masks generated with OPOM can generalize to different testing images to some degree. However, if there are obvious differences between the testing images and the training samples, such as, large poses, different illuminations, and occlusions, the mask protection tends to break down.}
	\label{fig:failurecase}
\end{figure}

\subsubsection{Protection against APIs}
In Figure~\ref{fig:API}, we conduct protection experiments against Commercial APIs (Amazon~\cite{Amazon}, Microsoft~\cite{azure}, Baidu~\cite{Baidu} and Face++~\cite{Face++}). Fifty identities in the Privacy-Commons dataset are randomly chosen for the face verification test, each with 5 test images. Since commercial APIs are based on extremely large datasets, a single model trained on CASIA-WebFace cannot obtain a better performance. With privacy masks generated with appropriate source models trained on CASIA-WebFace, VGGFace2~\cite{cao2018vggface2}, and MS-Celeb-1M~\cite{Guo16MS}, the average similarity score of the same person decreases significantly, which indicates better privacy protection. 

\subsection{Discussion}
\subsubsection{Failure Case Analysis}
Considering the existing performance of the person-specific universal privacy masks, it is pertinent to discuss the current challenges for privacy protection. We select some representative failure cases for analysis, as well as the corresponding training samples and successful easily protected samples, as shown in figure~\ref{fig:failurecase}, where each row represents an identity. We can see that, the generated privacy masks can generalize to different testing images to some degree. However, if there are obvious differences between the testing images and the training samples, such as  large poses, different illuminations and occlusions, the mask protection tends to break down. 

\subsubsection{Why person-spercific? Effectiveness and Efficiency} 
\begin{figure}[htbp]
	\center
	\includegraphics[width=0.98\linewidth]{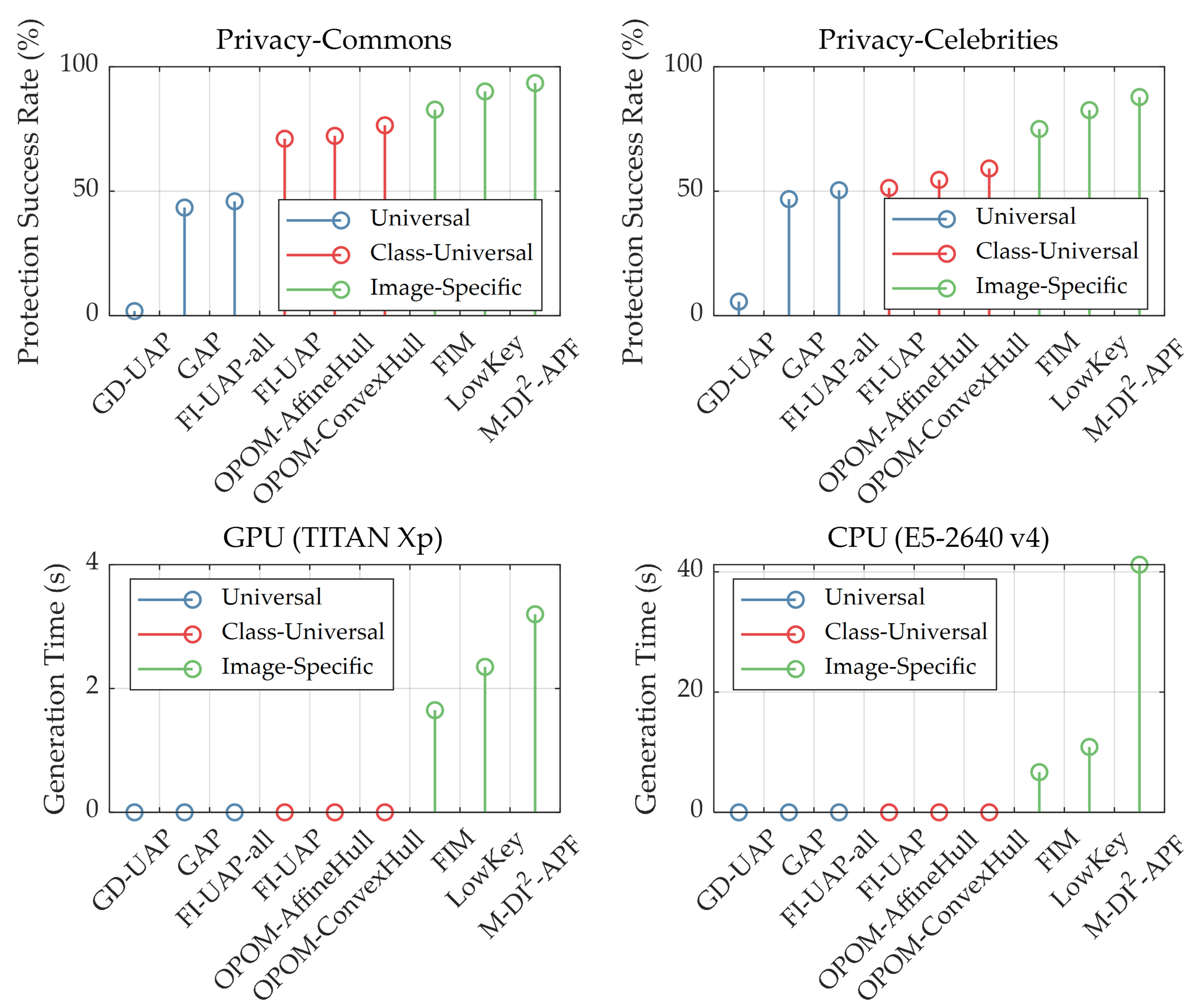}
	\caption{Comparison of universal, person-specific (class-wise) and image-specific masks in terms of effectiveness and efficiency.}
	\label{fig:eff}
\end{figure}

One might naturally wonder what are the strengths and weaknesses of person-specific (class-wise) universal masks compared with image-specific and universal masks. Here, we give an analysis in terms of effectiveness and efficiency. Specifically, universal masks (GD-UAP, GAP~\cite{poursaeed2018generative} and FI-UAP-all), person-specific universal masks (FI-UAP, OPOM-AffineHull and OPOM-ConvexHull), and image-specific masks (FIM~\cite{zhong2020towards}, LowKey~\cite{cherepanova2020lowkey} and M-DI$^{2}$-APF~\cite{zhang2020adversarial,dong2018boosting,xie2019improving}) are compared in Figure~\ref{fig:eff}. For brevity, we use the average protection success rate of six black-box models to represent the effectiveness, and use the average generation time of 100 images based on the ResNet-50 model to represent the efficiency. 

The experimental results show that the protection success rate of image-specific masks (FIM, LowKey and M-DI$^{2}$-APF) is the best, and the effectiveness of person-specific universal masks (FI-UAP, OPOM-AffineHull and OPOM-ConvexHull) is better than that of universal masks (GD-UAP, GAP and FI-UAP-all). While considering the efficiency, it takes $1.65s$, $2.35s$ or $3.50s$ to generate a mask of a new image for image-specific privacy protection, even with a GPU ($6.69s$, $10.88s$, and $41.20s$ with CPU). In contrast, person-specific universal masks and universal masks can dispense with the mask generation time for new images, that is, the generation time of new images is $0s$. Compared with universal masks and image-specific masks, person-specific (class-wise) universal masks show a tradeoff between effectiveness and efficiency, which may benefit average users and some real-time video applications.

\begin{figure*}[htbp]
	\center
	\includegraphics[width=1\linewidth]{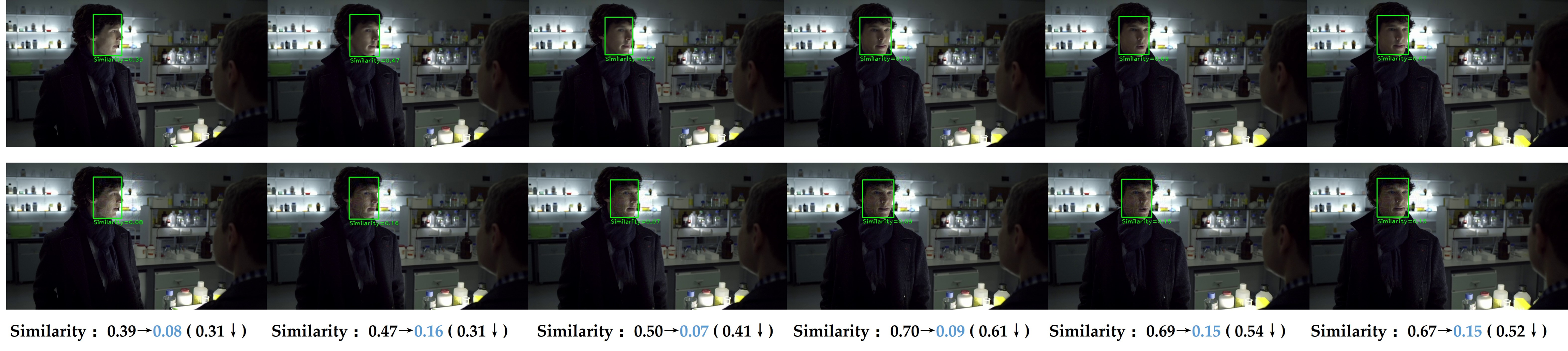}
	\caption{Application of OPOM in video privacy protection, Sherlock~\cite{Nagrani17b}. The first row shows the original frames. The second row shows the modified frames with the privacy masks of Sherlock generated from other images of actors (Benedict Cumberbatch). The cosine similarity between the deep features of the detected face in the video and the deep features of the corresponding character (Sherlock Holmes) is used to show the effectiveness. }
	\label{fig:shelock_video}
\end{figure*}

\begin{figure}[htbp]
	\center
	\includegraphics[width=1\linewidth]{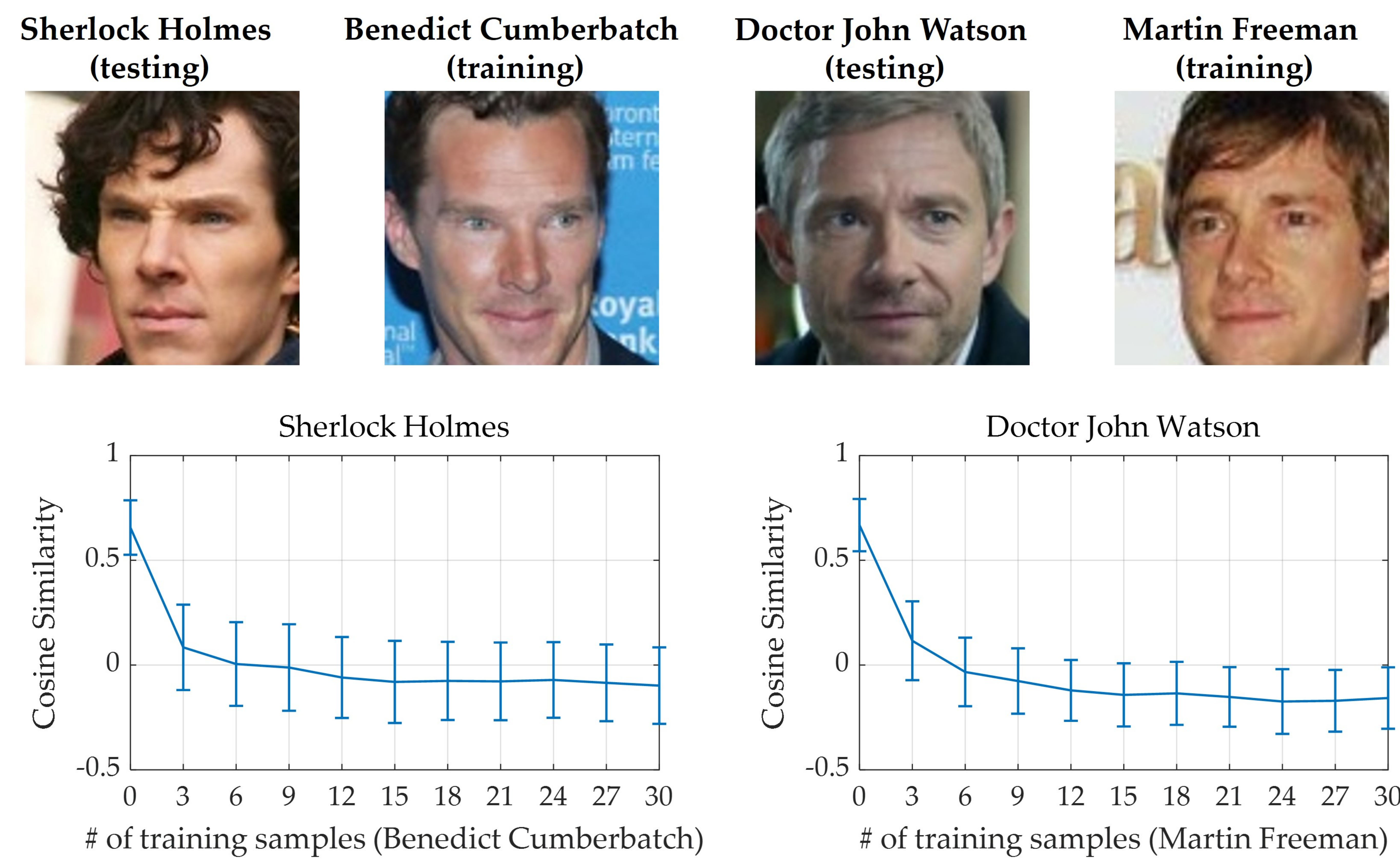}
	\caption{Application of OPOM in video privacy protection, Sherlock~\cite{Nagrani17b}. The privacy masks are trained with other face images of actors (Benedict Cumberbatch and Martin Freeman). The average cosine similarity between the deep features of the detected face in the video and the deep features of the corresponding characters (Sherlock Holmes and Doctor John Watson) is used to demonstrate the effectiveness. }
	\label{fig:shelock_sta}
\end{figure}

\begin{table*}[]
	\center
	\renewcommand\arraystretch{1.0}
	\caption{Diverse person-specific privacy masks ($\varepsilon = 8$) from a single source model to protect face images against black-box models. Here, ``M-O'' denotes that the masked image is used as the probe, while the original image is recorded in the gallery set. ``M-M'' represents masked image has been collected and applied in the gallery set, and new masked image is used as the probe. (The same mask is used for $n_M$=1.) We report the Top-1 protection success rate (\%) under 1:N identification setting of Privacy-Commons dataset. The higher protection success rate is better.}
	\label{table:table_diverse}
	\scalebox{0.925}{
		\begin{tabular}{@{}c|c|llllllllllll@{}}
			\toprule
			\multirow{3}{*}{\textbf{Source}} & \multirow{3}{*}{\textbf{${n_M}$}} & \multicolumn{12}{c}{\textbf{Target }}                                                                                                                                                                                                                                                                                                          \\
			&                                  & \multicolumn{2}{c|}{\textbf{ArcFace}}                    & \multicolumn{2}{c|}{\textbf{CosFace}}                    & \multicolumn{2}{c|}{\textbf{SFace}}                      & \multicolumn{2}{c|}{\textbf{MobileNet}}                  & \multicolumn{2}{c|}{\textbf{SENet}}                      & \multicolumn{2}{c}{\textbf{Inception-ResNet}} \\
			& & \textbf{M-O}   & \multicolumn{1}{c|}{\textbf{M-M}}   & \textbf{M-O}   & \multicolumn{1}{c|}{\textbf{M-M}}   & \textbf{M-O}   & \multicolumn{1}{c|}{\textbf{M-M}}   & \textbf{M-O}   & \multicolumn{1}{c|}{\textbf{M-M}}   & \textbf{M-O}   & \multicolumn{1}{c|}{\textbf{M-M}}   & \textbf{M-O}        & \textbf{M-M}        \\ \midrule
			
			\multirow{3}{*}{Softmax}                
			
			& 1  & 
			78.0 & 20.1 & 70.2 & 9.3 & 76.1 & 11.0 & 79.2 & 18.7 & 82.9 & 20.8 & 58.7 & 14.0  \\ 
			
			& 5  & 65.8$\pm$1.9 & \multicolumn{1}{c|}{82.8$\pm$4.2} & 56.2$\pm$3.4 & \multicolumn{1}{c|}{72.3$\pm$6.5} & 63.0$\pm$3.1 & \multicolumn{1}{c|}{79.4$\pm$5.9} & 66.6$\pm$2.4 & \multicolumn{1}{c|}{77.7$\pm$4.0} & 71.0$\pm$2.3 & \multicolumn{1}{c|}{85.9$\pm$3.3} & 43.6$\pm$2.8    &  66.6$\pm$4.1      \\ 
			
			& 10  & 54.7$\pm$8.3 & \multicolumn{1}{c|}{81.4$\pm$3.8} & 43.8$\pm$9.5 & \multicolumn{1}{c|}{71.5$\pm$5.5} & 50.4$\pm$9.7 & \multicolumn{1}{c|}{77.7$\pm$5.0} & 55.4$\pm$8.7 & \multicolumn{1}{c|}{76.6$\pm$3.7} & 59.3$\pm$9.0 & \multicolumn{1}{c|}{83.8$\pm$3.3} & 33.2$\pm$7.5 & 64.3$\pm$4.2    \\ \midrule
			
			\multirow{3}{*}{ArcFace}                
			
			& 1  & 86.5 & \multicolumn{1}{l|}{24.5} & 76.8 & \multicolumn{1}{l|}{11.5} & 82.7 & \multicolumn{1}{l|}{14.2} & 70.5 & \multicolumn{1}{l|}{18.0} & 79.3 & \multicolumn{1}{l|}{20.7} & 63.2 & 16.3\\ 
			
			& 5  & 78.2$\pm$2.3 & \multicolumn{1}{c|}{89.9$\pm$3.2} & 65.0$\pm$3.1 & \multicolumn{1}{c|}{76.6$\pm$6.3} & 71.6$\pm$2.9 & \multicolumn{1}{c|}{84.3$\pm$4.6} & 58.4$\pm$3.1 & \multicolumn{1}{c|}{70.3$\pm$4.0} & 68.4$\pm$3.4 & \multicolumn{1}{c|}{84.0$\pm$3.2} & 49.4$\pm$3.4     &  71.1$\pm$3.9      \\ 
			
			& 10  & 69.0$\pm$7.6 & \multicolumn{1}{c|}{88.3$\pm$2.4} & 53.5$\pm$9.3 & \multicolumn{1}{c|}{76.2$\pm$4.5} & 60.6$\pm$9.2 & \multicolumn{1}{c|}{82.8$\pm$3.5} & 47.4$\pm$8.3 & \multicolumn{1}{c|}{69.6$\pm$3.2} & 57.4$\pm$9.2 & \multicolumn{1}{c|}{82.3$\pm$2.8} & 38.4$\pm$8.1 & 68.8$\pm$3.5    \\ \midrule
			
			\multirow{3}{*}{CosFace}                
			
			& 1  & 86.6 & \multicolumn{1}{l|}{22.3} & 79.5 & \multicolumn{1}{l|}{11.9} & 83.0 & \multicolumn{1}{l|}{13.3} & 62.3 & \multicolumn{1}{l|}{15.7} & 74.0 & \multicolumn{1}{l|}{18.7} & 56.8 & 13.4\\ 
			
			& 5  & 76.3$\pm$2.9 & \multicolumn{1}{c|}{90.2$\pm$3.0} & 66.9$\pm$3.5 & \multicolumn{1}{c|}{82.1$\pm$5.3} & 71.0$\pm$3.6 & \multicolumn{1}{c|}{85.7$\pm$4.8} & 50.2$\pm$3.6 & \multicolumn{1}{c|}{63.6$\pm$3.9} & 62.0$\pm$4.0 & \multicolumn{1}{c|}{79.6$\pm$3.5} & 42.3$\pm$3.9     &  64.6$\pm$4.0      \\ 
			
			& 10  & 66.3$\pm$8.2 & \multicolumn{1}{c|}{88.4$\pm$2.5} & 54.6$\pm$10.0 & \multicolumn{1}{c|}{80.3$\pm$3.9} & 58.9$\pm$9.8 & \multicolumn{1}{c|}{83.8$\pm$3.7} & 39.9$\pm$8.0 & \multicolumn{1}{c|}{62.7$\pm$3.5} & 49.9$\pm$9.2 & \multicolumn{1}{c|}{77.8$\pm$3.3} & 31.7$\pm$7.8 & 61.9$\pm$3.7    \\ \midrule
	\end{tabular}}
\end{table*}

\subsubsection{Privacy Protection in Videos}
The proposed OPOM method can be applied to the privacy protection of photographs to be posted on the social media platform. The still image, of course, is not the only application. We demonstrate that OPOM is also applicable to the privacy protection in videos. 

Specifically, we use the series Sherlock database~\cite{Nagrani17b} for experiments. Following the original work~\cite{Nagrani17b}, we also collected images of actors that were the main characters by searching the names Benedict Cumberbatch and Martin Freeman. We manually found 10 high quality web-collected images of the two characters and use the mean value of their deep features to represent their feature subspace. Then, some other web-collected images, which can differ from the target images (from the TV material) in hairstyle, makeup, lighting, and viewpoint, are used as the training samples. Finally, we apply the generated privacy masks to target images of the videos. Some frames are shown in Figure~\ref{fig:shelock_video}, where the first row shows the original frames of the video, and the second row shows the modified frames with the privacy masks of Sherlock generated from other face images of actors (Benedict Cumberbatch). 

The effectiveness of the privacy protection is evaluated using the cosine similarity between the deep features of images from videos and the feature prototype. As shown in Figure~\ref{fig:shelock_sta}, the average value and the standard deviation of cosine similarity between the face features of characters (Sherlock Holmes and Doctor John Watson), tested on six black-box models, are used to demonstrate the effectiveness of OPOM. With no privacy masks (no training images) provided, it is easy to recognize the character. With privacy masks, privacy can be protected against black-box deep face recognition models. We gradually increase the number of web images for training and found that 15 training images can obtain good protection performance. 

\subsubsection{Diversity of Privacy Masks}
In this paper, we apply privacy masks to the probe images, not the gallery images, since we assume that some unauthorized face recognition services may have already obtained the original images of regular users. Despite the effectiveness of person-specific (class-wise) universal masks in one-shot usage, we have to consider a more difficult situation since new images on the internet are being continuously collected. If masked face images have been labeled and recorded in the gallery set, then new images with the same mask (probe) may no longer be safe. Therefore, we explore whether a diversity of person-specific masks can be generated for an identity to advance the possibility of potential solutions for this problem. 

That is, the objective is to generate a set of diverse person-specific (class-wise) universal masks
to fool a variety of deep face recognition models $f(\cdot)$ for all the face images $ {{X}^k} =\left\{ X_1^k,X_2^k,\ldots ,X_i^k, \ldots \right\}$ of identity $k$, so that any face image $X_{i}^{k}$ can conceal the identity with any mask of $\left\{ \Delta X_1, \Delta X_2, \cdots , \Delta X_j, \cdots \right\}$:
\begin{equation} 
	\label{equ:obj_diverse}
	\begin{gathered}
		D( f( X_{i}^{k}+\Delta {X_{j_1}}), f_{{{X}^k}}) > t, \hfill \\
		D( f( X_{i}^{k}+\Delta {X_{j_2}}), f_{{{X}^k}}) > t, \hfill \\
		D( f( X_{i}^{k}+\Delta {X_{j_1}}), f( X_{i}^{k}+\Delta {X_{j_2}}) ) > t, \hfill\\ 
		\qquad\qquad\qquad\qquad\qquad\qquad\lVert \Delta {X_{j_1}} \rVert _{\infty}<\varepsilon, \lVert \Delta {X_{j_2}} \rVert _{\infty}<\varepsilon. 
	\end{gathered}
\end{equation}In the above, $f(X_{i}^{k})\in \mathbb{R}^d$ is the normalized feature of image $X_{i}^{k}$. $f_{{{X}^k}}$ denotes the feature subspace of identity $k$. $\Delta {X_{j_1}}$ and $\Delta {X_{j_2}}$ are any two generated privacy masks of the set. $t$ is the distance threshold to decide whether a pair of face images belongs to the same identity. $D(x_1,x_2)$ denotes the distance between $x_1$ and $x_2$, that is, the shortest distance between point $x_1$ and subspace $x_2$. When $x_2$ denotes a point, we use $D(x_1,x_2)$ to represent the normalized Euclidean distance or cosine distance commonly used in face recognition. $\varepsilon$ limits the maximum deviation of the privacy mask. 

The objective not only optimizes each masked sample in the direction away from the feature subspace of the source identity, but also optimizes samples with different masks away from each other. 

Considering the above function, we still use the proposed approximation methods for $f_{{{X}^k}}$ with a limited number ${n_k}$ of face images $\tilde{{{X}^k}} = \left\{ X_1^k,X_2^k,\ldots ,X_{n_k}^k\right\}$ and generate $n_M$ masks $\left\{ \Delta X_1, \Delta X_2, \cdots , \Delta X_{n_M}\right\}$ with previous source models. For experimental settings, in addition to the original image as gallery set, we also consider that different masked images are used in the gallery set. 

Experimental results in Table~\ref{table:table_diverse} indicate that person-specific privacy masks can be diverse and varied. In this way, the average user can apply different masks for privacy protection each time to cope with the situation that some masks have been ineffective. However, it seems that the protection rate will be lessened as the number of privacy masks increases. Note that we are only presenting a preliminary exploration; it will be for future work to study more elaborate solutions for this purpose.

\section{Conclusion and Future Works}
\label{sec:conclusion}
In this paper, we have presented a type of class-wise universal adversarial perturbation for a new customized privacy protection task, by generating a person-specific masks that can be applied universally for all the images of an identity. Experimental results on two test benchmarks have demonstrated the effectiveness and superiority of the proposed method on this customized privacy protection task. We have demonstrated that the proposed method can be used in the privacy protection of videos. 

Great progress has been made, yet much still remains to be done. As we stated before, it is challenging to offer protection if there are obvious differences between the testing images and the training samples. Therefore, it is imperative to further improve the image universality to cover the potential diverse variances of face images, such as large poses, different illuminations, and occlusions. As analyzed before, we considered a diverse of masks to address the potential problems of privacy mask leakage. However, the proposed method has challenges in terms of the diversity and effects of the privacy masks, and therefore remains a preliminary exploration. In addition, we are currently at the implementation stage of generating digital privacy cloaks that can only be used for digital images and videos on social media. They will have more impact if they can be applied in practical-world video surveillance.  

\section*{Acknowledgment}
We would like to thank Jinglin Zhang and Xuannan Liu for useful feedback, and anonymous referees for their valuable comments. This work was partially supported by the National Natural Science Foundation of China under Grants No. 61871052 and 62192784.

\ifCLASSOPTIONcaptionsoff
  \newpage
\fi



%
\bibliographystyle{IEEEtran}
\bibliography{privacy}

%
%




\end{document}